\documentclass[symmetry,article,accept,oneauthor]{Definitions/mdpi}
\AtBeginDocument{\let\linenumbers\relax}
\makeatletter
\renewcommand{\contentleftcolumn}{}
\patchcmd{\@maketitle}{\includegraphics[height=1.2cm]{Definitions/\@journal-logo.eps}}{}{}{\typeout{WARN: journal-logo patch failed}}
\patchcmd{\@maketitle}{\includegraphics[height=1cm]{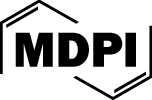}}{}{}{\typeout{WARN: mdpi-logo patch 1 failed}}
\patchcmd{\@maketitle}{\includegraphics[height=1cm]{Definitions/logo-mdpi.eps}}{}{}{\typeout{WARN: mdpi-logo patch 2 failed}}
\makeatother
\AtBeginDocument{%
  \lhead{}%
  \rfoot{}\cfoot{}%
  \fancypagestyle{plain}{\fancyhf{}}%
}

\graphicspath{{figures/}}
\usepackage{newtxtt}

\firstpage{1} 
\pubvolume{1}
\issuenum{1}
\articlenumber{0}
\pubyear{2026}
\copyrightyear{2026}
\externaleditor{Firstname Lastname} %
\datereceived{9 July 2026} 
\daterevised{30 July 2026} 
\dateaccepted{2 August 2026} 
\datepublished{ } 

\Title{Classical Versus Deep Mirror-Symmetry Scoring: A Benchmark of Thirteen Methods}

\Author{Maximilian Woehrer $^{1,2,}$\orcidA{}} %

\address{%
$^{1}$ \quad Research Group Software Architecture, Faculty of Computer Science, University of Vienna, \mbox{1090 Vienna, Austria}; maximilian.woehrer@univie.ac.at\\%

$^{2}$ \quad MAXI solutions e.U., 1170 Vienna, Austria
}

\abstract{Quantifying how mirror-symmetric an image is about a given axis (\emph{symmetry scoring}) underpins applications from visual aesthetics to medical imaging, yet proposed scoring methods have never been compared on a common, statistically grounded protocol. We benchmark 13 scoring methods (9 collected from the literature; 4 introduced here) spanning from classical features to frozen deep features, across four single-axis and five multi-axis datasets under a reflection-exact protocol with a chance-anchored, significance-tested discrimination skill.
Deep backbones perform best on single-axis and harder multi-axis protocols. 
However, a classical histogram-of-oriented-gradients (HOG) descriptor trails the best frozen-network readout by a small (but significant) margin, is not statistically separable from the runner-up (a CNN-filter measure), and runs ${\sim}300\times$ faster on CPU.
Our results show that discrimination concentrates in mid-scale oriented features, where deep backbones peak at a low or mid stage, and HOG peaks at a mid cell size.
Among existing methods, frozen deep features thus offer little over a tuned classical descriptor for measuring symmetry; whether task-trained deep scorers can do better remains open. We release the scorers and harness, in \texttt{imgsym}, an open toolkit for image symmetry detection and measurement.} %



\begin{document}

\section{Introduction}\label{sec:intro} %

Bilateral (mirror) symmetry is one of the most striking patterns in natural and man-made images and can be found in many contexts, such as faces, animals, architecture, and patterns. The human visual system
quickly detects symmetries and uses them as indicators of salience, figure-ground organization, and aesthetic
preference~\cite{reisfeldDetectionInterestPoints1990,bonneh2003quantification,funkSymmetryReCAPTCHA2016}.
Turning that percept into a number (a scalar quantifying how symmetric an image is about a
given axis, or equivalently how much it departs from symmetry) underpins applications from the
quantification of anatomical asymmetry in medical
imaging~\cite{smithAccurateRobustSymmetry1999,hogewegFastEffectiveQuantification2017} and of
developmental asymmetry in biology~\cite{freyMethodQuantifyingRotational2007} to visual
aesthetics and
design~\cite{mayerQuantifyingVisualAesthetics2018,gartusPredictingPerceivedVisual2017}.

Two distinct problems are often conflated under ``symmetry''. \emph{Detection} localizes the %
symmetry axis (or axes) in an image; \emph{scoring} (or discrimination) quantifies how
symmetric the content is about a given axis. The two are not interchangeable: a detector
answers \emph{where} a symmetry is, whereas a scorer answers \emph{how much} symmetry there is
about a specified axis, which is the quantity the applications above actually consume. This
paper concerns symmetry scoring.

Throughout many fields a variety of scoring methods have been proposed, mostly crafted for a specific purpose and targeting a specific application or dataset. 
They range from hand-crafted measures built on gradients~\cite{gnuttiCombiningAppearanceGradient2021}, filter banks~\cite{shakerNewSymmetryMeasure2015}, frequency coefficients~\cite{gunluSymmetryAnalysis2D2009}, or pixel correlations~\cite{mayerQuantifyingVisualAesthetics2018} to more recent measures read from deep-network features~\cite{brachmannUsingConvolutionalNeural2016}.
Yet for all this variety, symmetry-scoring methods have never been compared head to head on
the same everyday images, under a common protocol with significance testing. The primary question of this paper is therefore
practical: \emph{which of the proposed methods measures mirror symmetry well---and at what
computational cost?} Answering it also addresses, for the methods that exist today, a second
question that the field's drift toward learned features keeps raising: whether reading
symmetry from frozen pretrained features earns its cost over classical descriptors.

{Our contributions are as follows:} %
\begin{enumerate}
\item \textbf{The first benchmark of symmetry scoring}, with a unified open library (\texttt{imgsym}) of its %
methods. We recast 13 scorers (9 from three decades of scattered literature; 4 introduced
here) as instances of a single representation--comparison--aggregation template
(Equation~\eqref{eq:template}) and evaluate them on a reflection-exact harness with a
chance-anchored, paired-bootstrap protocol across four single-axis and five multi-axis datasets
(${\sim}3950$ axis units; Sections~\ref{sec:methods} and~\ref{sec:scoremethods}).
\item \textbf{Discrimination lives in mid-scale, unsigned, oriented features.} Skill peaks at a
low or mid deep-network stage and at a mid HOG cell size (two independent routes), unsigned
orientation beats signed, and columnar Vision Transformers perform poorly at standard $16$-px
tokenization but partly recover with a finer-patch control, pointing to access to mid-scale
spatial structure rather than depth hierarchy or capacity
(Section~\ref{sec:principle}).
\item \textbf{An actionable cost--skill map.} A tuned classical HOG trails the best
frozen-feature readout by only $0.03$ skill (small but significant), is not statistically
separable from AlexNet-C2, and runs ${\sim}300\times$ faster on a CPU; whether task-trained
deep scorers can widen that gap is an open question (Sections~\ref{sec:leaderboard}
and~\ref{sec:speed}).
\end{enumerate}

After %
the related work (Section~\ref{sec:related}), Section~\ref{sec:methods} defines
the benchmark (task, data, extraction, negatives, and metric), and Section~\ref{sec:scoremethods} %
defines the thirteen methods; Section~\ref{sec:results} reports the results as four questions, and
Sections~\ref{sec:discussion} and~\ref{sec:conclusions} %
offer interpretations and the conclusion.

\section{Related Work}\label{sec:related}

Computational symmetry spans two broad problems, \emph{detecting} symmetry axes and
\emph{scoring} symmetry about a given axis. We review both below, along with the competitions
and datasets used to evaluate them.

\subsection{Symmetry Detection}\label{sec:rel-detection}
The larger body of work localizes symmetry axes; we give only a coarse map here (fuller
algorithm surveys appear in the competition
reports~\cite{liuSymmetryDetectionRealWorld2013,funk2017ICCVChallenge2017a}), and the cleanest
division is by input. Methods for \emph{images} operate on gray-level or color pixels, %
through
mirrored-feature correspondence and
voting~\cite{loyDetectingSymmetrySymmetric2006,choBilateralSymmetryDetection,caiAdaptiveSymmetryDetection2014,wangReflectionSymmetryDetection2015};
dense per-axis evidence from edges, orientations, wavelets, or
phase~\cite{sunFastReflectionalSymmetry1999,cicconetConvolutionalApproachReflection2016,elawadyWaveletBasedReflectionSymmetry2017,guerriniImageSymmetriesRight2017,xiaoUsingPhaseInformation2005};
registration of the image against its own
reflection~\cite{cicconetFindingMirrorSymmetry2017}; and, recently, learned axis
prediction~\cite{tsogkasLearningBasedSymmetryDetection2012,keSRNSideOutputResidual2017,seoReflectionRotationSymmetry2022,yuAxislevelSymmetryDetection2025,yangCLIPSymDelvingSymmetry2025}.
Methods geared toward \emph{binary-shape analysis} assume a segmented silhouette or mask and
score candidate axes with projection, boundary, or skeleton
signatures~\cite{nguyenProjectionBasedApproach2019,nguyenReflectionSymmetryDetection2022,lomovDetectionOptimalReflection2022,mestetskiyMirrorSymmetryDetection2020,kushnirALGORITHMSADJUSTMENTSYMMETRY2019};
they presuppose a segmentation and do not transfer to unsegmented images. Orthogonal to this
division, a contrario techniques contribute a statistical \emph{validation} stage rather than
a detection family, accepting a candidate symmetric patch or match only when its expected
number of false alarms is
low~\cite{patrauceanDetectionMirrorSymmetricImage2013,vongioiContrarioPatchMatching2015}; and
a classical line scores \emph{local point symmetry} to find interest points or saliency maps
rather than global
axes~\cite{reisfeldDetectionInterestPoints1990,dalitzGradientProductTransform2019}.

The current state of the art is \emph{equivariant}: networks whose architecture respects
reflection. EquiSym predicts dense axis heatmaps with reflection-equivariant
convolutions~\cite{seoReflectionRotationSymmetry2022}, CLIPSym pairs a vision--language
backbone with a rotation--reflection-equivariant decoder~\cite{yangCLIPSymDelvingSymmetry2025},
and axis-level detection drops the heatmap altogether, predicting each axis directly as a
geometric primitive (a line segment) from dihedral-group-equivariant
features~\cite{yuAxislevelSymmetryDetection2025}.
These models are trained end to end for detection: they emit spatial axis evidence, not a
scalar score for a given axis, and turning their dense outputs into a per-axis score would
require an aggregation choice that would itself define a new, task-trained scorer, which no
published work provides; we return to task-trained scoring as an open direction
(Section~\ref{sec:discussion}). We do not
benchmark detectors as scorers (Section~\ref{sec:task} formalizes how the two tasks relate)
and instead reuse their annotated data, including DENDI, the dataset EquiSym introduced
(Section~\ref{sec:datasets}).

\subsection{Symmetry Scoring}\label{sec:rel-scoring}
A smaller, more scattered literature assigns a scalar symmetry score to a given axis.
We order it from low-level to high-level representations, the same progression as our
benchmark pool (the 13 methods we evaluate; Section~\ref{sec:scoremethods}). At the pixel level, perceptual studies score
weighted mirrored-pixel agreement on abstract binary (black-and-white) patterns to predict
human
ratings~\cite{bauerlyComputationalModelingExperimental2006,hubnerComparisonObjectiveMeasures2016,gartusPredictingPerceivedVisual2017},
and aesthetics research applies pixel-luminance reflection correlation to
photographs~\cite{mayerQuantifyingVisualAesthetics2018}. Whether a method assumes a binary
mask or a gray-level image is a divide that recurs throughout this literature. The asymmetry
indices developed in application domains build on the same pixel-intensity comparisons:
mid-sagittal-plane
estimation in brain imaging, pathology screening on chest radiographs, and floral asymmetry
in developmental
biology~\cite{smithAccurateRobustSymmetry1999,liuRobustMidsagittalPlane2000,hogewegFastEffectiveQuantification2017,freyMethodQuantifyingRotational2007}.

Mid-level measures score the symmetry of image structure: DCT parity, paired Gabor
orientations, and HOG mirror correlation, developed for face analysis and painting
aesthetics~\cite{gunluSymmetryAnalysis2D2009,shakerNewSymmetryMeasure2015,renero-cExtractingSymmetryHuman2017};
several detection pipelines likewise embed per-axis merit functions (even/odd energy
balances~\cite{guerriniImageSymmetriesRight2017}; appearance-plus-gradient
agreement~\cite{gnuttiCombiningAppearanceGradient2021}) that
are effectively per-axis scores, and three benchmark methods formalize such criteria as
standalone scorers.

At the top of the progression, a deep line measures the symmetry of pretrained CNN
filter responses~\cite{brachmannUsingConvolutionalNeural2016,rediesToolboxCalculatingQuantitative2025};
the Aesthetics Toolbox~\cite{rediesToolboxCalculatingQuantitative2025} unifies several of these
measures in one implementation but compares none of them on task performance.
Scalar scores also appear at other granularities (symmetric-SSIM quality
indices~\cite{cuiImageQualityMetric2008}, per-pixel symmetry
maps~\cite{nagarSymmMapEstimation2D2017}, and boundary-Fourier asymmetry of segmented
shapes~\cite{mestetskiyMirrorSymmetryDetection2020}), none of which enters the pool (a quality
index needs a reference image; per-pixel maps answer a local question; shape scores need a
mask), while the binary-mask line is represented through its variant WBS, which binarizes
internally. These measures were proposed independently, for different tasks and data, and were
never compared on a common, chance-anchored protocol with significance
testing: the gap this benchmark closes. Table~\ref{tab:landscape} organizes the families,
their native tasks, and the benchmark methods each contributes.

\begin{table}[H]
\caption{An outline of the symmetry-scoring literature: method families with their native
inputs and tasks, representative (not exhaustive) references, and the benchmark methods
(Section~\ref{sec:scoremethods}) each family~contributes.}\label{tab:landscape}
\begin{adjustwidth}{-\extralength}{0cm}
\begin{tabularx}{\fulllength}{>{\raggedright\arraybackslash\hsize=0.9\hsize}Xl>{\raggedright\arraybackslash\hsize=1.1\hsize}Xl}
\toprule
\textbf{Family} & \textbf{Native Input} & \textbf{Native Task} & \textbf{In Benchmark Pool}\\
\midrule
Perceptual pattern measures~\cite{bauerlyComputationalModelingExperimental2006,hubnerComparisonObjectiveMeasures2016,gartusPredictingPerceivedVisual2017} & binary patterns & predicting human symmetry and complexity ratings via weighted mirrored-pixel agreement & WBS\\

\bottomrule
\end{tabularx} 
\end{adjustwidth}
\end{table}

\begin{table}[H]\ContinuedFloat
\caption{\textit{Cont.}}
\begin{adjustwidth}{-\extralength}{0cm}
\begin{tabularx}{\fulllength}{>{\raggedright\arraybackslash\hsize=0.9\hsize}Xl>{\raggedright\arraybackslash\hsize=1.1\hsize}Xl}
\toprule
\textbf{Family} & \textbf{Native Input} & \textbf{Native Task} & \textbf{In Benchmark Pool}\\
\midrule
Aesthetics measures~\cite{mayerQuantifyingVisualAesthetics2018} & images & predicting aesthetic liking via pixel-luminance reflection correlation & PixCorr\\
\midrule
Domain asymmetry indices~\cite{smithAccurateRobustSymmetry1999,liuRobustMidsagittalPlane2000,hogewegFastEffectiveQuantification2017,freyMethodQuantifyingRotational2007} & images & brain mid-sagittal-plane estimation; pathology screening on chest radiographs; floral developmental asymmetry & EROS, PatchNN\\
\midrule
Spectral \& oriented measures~\cite{gunluSymmetryAnalysis2D2009,shakerNewSymmetryMeasure2015,renero-cExtractingSymmetryHuman2017} & images & face and painting symmetry via DCT parity; paired Gabor orientations; HOG mirror correlation & DCT, Gabor, HOG, PHOG\\
\midrule
Detection-embedded merits~\cite{guerriniImageSymmetriesRight2017,gnuttiCombiningAppearanceGradient2021} & images & ranking candidate axes inside detection pipelines via even/odd energy balance or appearance-plus-gradient agreement & SlideWin, Grad, MS-Grad\\
\midrule
CNN-feature measures~\cite{brachmannUsingConvolutionalNeural2016,rediesToolboxCalculatingQuantitative2025} & images & predicting perceived symmetry and aesthetics from pretrained filter responses & AlexNet-C2, DeepFeat\\
\midrule
Similarity \& shape scores~\cite{cuiImageQualityMetric2008,nagarSymmMapEstimation2D2017,mestetskiyMirrorSymmetryDetection2020} & masks/images & image-quality assessment; per-pixel symmetry mapping; silhouette shape analysis & ---(see text)\\
\bottomrule
\end{tabularx}
\end{adjustwidth}
\end{table}

\subsection{Competitions and Datasets}\label{sec:rel-data}
Symmetry has been benchmarked mainly through detection competitions that score the precision and
recall of predicted axes against human ground
truth~\cite{liuSymmetryDetectionRealWorld2013,funk2017ICCVChallenge2017a}. Their released image
sets, together with more recent dense and perceptually graded
collections~\cite{cicconetConvolutionalApproachReflection2016,seoReflectionRotationSymmetry2022,muradasodriozolaPixelsPerceptionBenchmark2026},
supply the annotated axes that we repurpose for discrimination (Section~\ref{sec:datasets}). We adopt
their images but not their localization metric: instead of scoring whether an axis is
found, we score how well a given axis is ranked against controlled perturbations of it.

\section{The Benchmark: Task, Data, and Protocol}\label{sec:methods}

\subsection{Task: Symmetry Discrimination}\label{sec:task}
\begin{figure}[H]
\includegraphics[width=\textwidth]{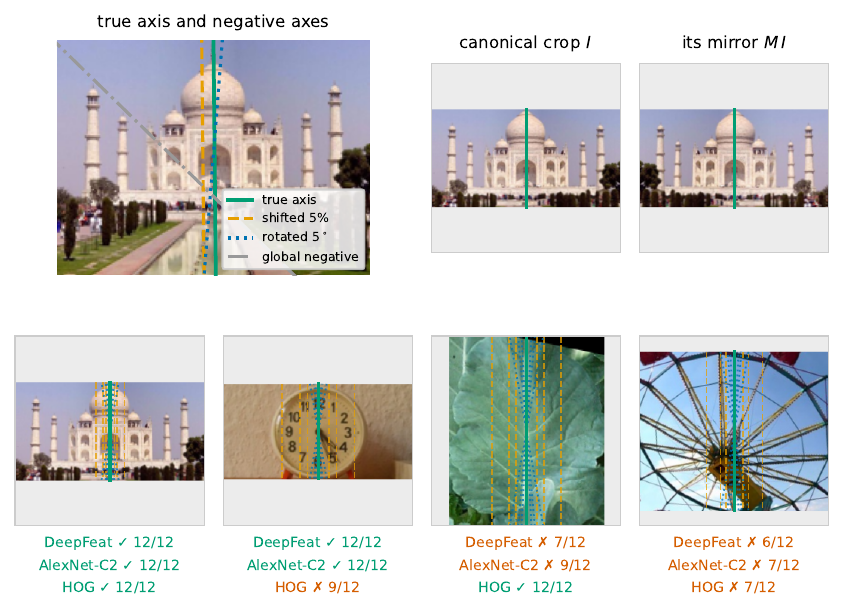}
\caption{The discrimination task in practice. Top: an input image with its true axis and one
negative axis of each kind (Section~\ref{sec:perturb}), and the canonical crop pair the scorer
receives: the whole crop $I$ and its mirror $M\,I$ (Section~\ref{sec:extraction}). Bottom:
four benchmark crops spanning the outcome range; thin lines mark each crop's 12 standard
perturbed negatives (orange = shifted; blue = rotated) around the true axis, and the verdicts
give how many of them the true axis outranked (\checkmark{} = all 12; $\times$ = fewer than all 12, with the count shown).
\label{fig:task}} %
\end{figure}

A \emph{symmetry scorer} maps an image and a candidate reflection axis to a scalar that should
increase with the strength of mirror symmetry about that axis. Our harness puts the
geometry in a canonical form (Section~\ref{sec:extraction}): every candidate axis is mapped onto the exact vertical
centerline of an extracted crop, so each scorer only ever judges the left--right symmetry of a
canonical image $I$, that is, how well $I$ agrees with its mirror $M\,I$ ($M$ denotes the
left--right mirror throughout). The benchmark task is
\emph{discrimination}: for each annotated true axis and each wrong axis
(Section~\ref{sec:perturb}), a scorer succeeds when it assigns the true axis's crop a higher
score than the wrong axis's crop. Figure~\ref{fig:task} illustrates the task and what success
and failure look like.

Two dimensions are deliberately not properties of a scorer. First, axis \emph{search}: a
scorer takes the axis as an input; detection (localizing axes in an
image) can be viewed as a search over candidate axes wrapped around a scorer, or approached by
dedicated detectors~\cite{loyDetectingSymmetrySymmetric2006,cicconetFindingMirrorSymmetry2017},
and is not benchmarked here. Second, \emph{invariances}: robustness to blur, occlusion, or
rescaling is a property one probes of a scorer with benchmark perturbations, not an
intrinsic attribute of the method; it therefore does not enter the taxonomy of
Section~\ref{sec:scoremethods}.

\subsection{Datasets and Protocols}\label{sec:datasets}
The available datasets come in two annotation styles (some annotate a single dominant
reflection axis per image, others annotate every visible reflection axis), so we evaluate
under two matching protocols (Table~\ref{tab:datasets}). The \emph{single-axis} protocol
scores one dominant annotated axis per image across four datasets drawn from the symmetry
competitions and perception benchmarks of Section~\ref{sec:rel-data}. The \emph{multi-axis}
(per-axis) protocol treats every annotated reflection axis as one discrimination unit (one unit
per image--axis pair) across five further sets, adding the modern, in-the-wild DENse and DIverse symmetry dataset (DENDI)~\cite{seoReflectionRotationSymmetry2022}.

The single-axis sets are as follows: \emph{symComp17\_s}, the single-reflection training images of the
ICCV~2017 symmetry-detection competition~\cite{funk2017ICCVChallenge2017a} (we evaluate on the
training split because the competition's test ground truth was never publicly released);
\emph{NYU\_s}, the single-axis subset of the NYU symmetry
database~\cite{cicconetConvolutionalApproachReflection2016}; and \emph{PIX2PER-nat} and
\emph{PIX2PER-art}, the natural-photograph and artwork halves of the crowd-annotated PIX2PER
benchmark~\cite{muradasodriozolaPixelsPerceptionBenchmark2026}, for which we take each image's
most frequently annotated axis as its dominant axis. The CVPR~2013 competition's
single-reflection set~\cite{liuSymmetryDetectionRealWorld2013} is excluded: 25 of its 35
images recur in symComp17\_s with matching ground truth (a verified competition-lineage
overlap), and its 10 unique images are too few to stand as an independent
dataset. The multi-axis sets reuse the multiple-reflection portions of symComp13 (a subset
disjoint from its single-reflection images, with no symComp17 counterpart) and NYU, score
all annotated PIX2PER axes, and add the COCO-image reflection subset of
DENDI~\cite{seoReflectionRotationSymmetry2022} (introduced with the EquiSym detector). We take
DENDI's reflection-axis annotations on COCO photographs only, disjoint from the NYU and
symComp images that DENDI's separate polygon-annotation split re-labels. The multi-axis
protocol alone contributes ${\sim}3300$ of the benchmark's ${\sim}3950$~axis~units.

\begin{table}[H]\small
\caption{Evaluation datasets. The single-axis protocol scores one dominant axis per image; the
multi-axis protocol scores every annotated reflection axis (one unit per image--axis pair);
images with more than 10 annotated axes are excluded (Section~\ref{sec:datasets}).\label{tab:datasets}}
\begin{tabularx}{\textwidth}{lccL}
\toprule
\textbf{Dataset} & \textbf{\# Images} & \textbf{\# Axis Units} & \textbf{Notes}\\
\midrule %
\multicolumn{4}{@{}l}{\emph{Single-axis protocol}}\\  %
symComp17\_s~\cite{funk2017ICCVChallenge2017a} & 99  & 99   & ICCV'17 competition, single reflection (train split)\\
NYU\_s~\cite{cicconetConvolutionalApproachReflection2016}       & 172 & 172  & NYU symmetry database, single-axis subset\\
PIX2PER-nat~\cite{muradasodriozolaPixelsPerceptionBenchmark2026}  & 198 & 198  & PIX2PER natural photographs (dominant axis)\\
PIX2PER-art~\cite{muradasodriozolaPixelsPerceptionBenchmark2026}  & 199 & 199  & PIX2PER artworks (dominant axis)\\
\midrule
\multicolumn{4}{@{}l}{\emph{Multi-axis protocol (every annotated axis; images with $>$10 axes excluded)}}\\
symComp13\_m~\cite{liuSymmetryDetectionRealWorld2013} & 16  & 39   & CVPR'13 competition, multiple-axis~set\\
NYU\_m~\cite{cicconetConvolutionalApproachReflection2016}       & 63  & 174  & NYU multiple-axis subset\\
DENDI~\cite{seoReflectionRotationSymmetry2022}        & 419 & 1630 & DENse and DIverse symmetry dataset, reflection subset (EquiSym)\\
PIX2PER-nat~\cite{muradasodriozolaPixelsPerceptionBenchmark2026}  & 200 & 690  & PIX2PER natural photographs (all axes)\\
PIX2PER-art~\cite{muradasodriozolaPixelsPerceptionBenchmark2026}  & 199 & 752  & PIX2PER artworks (all axes)\\
\bottomrule
\end{tabularx}
\end{table}

Two protocol details apply to the multi-axis runs. Because DENDI annotates no dominant axis and
some images contain dense, near-regular tilings, we score every annotated axis but drop images
with more than $K=10$ axes, which are ill-posed for per-axis discrimination. Because DENDI's
in-the-wild images are large, we also downscale any crop whose longest side exceeds $256$~px; the
single-axis sets are essentially unaffected, and the deep method resizes to $224$ internally
regardless. This cap is a tractability measure, not a result-changing one: on a fixed DENDI
subset, it leaves the leading order unchanged (including DeepFeat~$\approx$~HOG), produces only mid-pack swaps within overlapping intervals, and changes
per-method skill by less than $0.06$ (mean $0.02$), within the $95\%$ confidence intervals.

\subsection{Reflection-Exact Extraction}\label{sec:extraction}
Each image--axis pair is mapped to a canonical crop by a single warp that places
the axis on the exact vertical centerline: the crop has even width and is sampled with a
half-pixel offset so that flipping the crop left--right is an exact reflection about the
candidate axis, with no resampling asymmetry between the two halves. (The axis direction
convention determines which member of the pair $\{I, M\,I\}$ the warp returns: an upward-
versus downward-pointing axis yields mirrored twins; this is immaterial, since every
comparison in Section~\ref{sec:scoremethods} is symmetric in the pair.) Out-of-image regions are
filled with a constant border (never reflection padding, which would inject spurious
symmetry), and crops with too little image support are discarded. The crop
extent follows the annotation type: segment-annotated datasets extend to the nearer image edge,
and PIX2PER uses its annotated boxes. Single-axis runs are evaluated at native resolution; the
multi-axis size cap is described in Section~\ref{sec:datasets}.

One extraction detail requires a control. Under the default extent policy, the crop's
perpendicular width is bounded by the nearer image edge, so a shifted (wrong) axis receives a
systematically narrower crop than the true axis; crop geometry thereby correlates with
the correct answer, and a scorer sensitive to crop size could gain skill without measuring
symmetry. We therefore re-ran a representative subset of methods (the three leaders plus the
most size-suspect cheap measures) with fixed extents, giving the true axis and every
wrong axis of an image pixel-identical crop dimensions: the ordering is essentially unchanged
(the top two methods swap the lead, well within their confidence intervals), and per-method
skills change by less than $0.04$, comparable to those intervals.
Discrimination skill on this benchmark is therefore content-driven, not a crop-geometry
artifact.

\subsection{Negative Axes: Perturbed and Global}\label{sec:perturb}
Each true axis is paired with \emph{perturbed} negatives, copies of the true axis displaced
by a graded amount (one example of each kind is drawn in Figure~\ref{fig:task}): perpendicular
shifts of $\{0.5,1,2,3,5,10\}\%$ of the crop size and in-plane rotations of
$\{0.5,1,2,3,5,10\}^{\circ}$, each applied in both directions. Both kinds matter (shifts alone would
never probe orientation errors), and every method scores the identical negatives, which
licenses the paired statistics of Section~\ref{sec:metric}. The main results use only the
coarser half of these magnitudes (shifts $\geq 3\%$; rotations $\geq 3^{\circ}$): the finest
perturbations are near-duplicates of the true axis that no method can reliably tell apart, and
letting them dominate would compress all skills toward chance. How skill varies over the full
range of magnitudes is analyzed separately in Section~\ref{sec:leaderboard}.

Each image additionally receives 12 \emph{global} structural negatives (cardinal
center/offset lines and seeded-random lines), reported separately (Section~\ref{sec:localglobal});
any global negative that nearly coincides with the true axis (angle within $10^\circ$ and normal
offset within 5\% of the short side, common on centered subjects, where the vertical center line
can be the true axis) is excluded, so methods are never punished for scoring the truth
(5.6\% of global negatives across the four sets). A residual caveat remains: a global negative
can coincide with a genuine but unannotated symmetry (for example, a water reflection
about a near-horizontal center line), in which case a correct method is penalized for noticing
real symmetry. Because this deflates precisely the best scorers, the separately reported
global-rejection numbers (Section~\ref{sec:localglobal}) are conservative; the primary skill,
computed from perturbed negatives only, is unaffected. All negatives are deterministic: the
perturbation ladders are fixed, and the seeded-random global lines are generated per image
from its dataset index, so every run scores the identical negative set.

\subsection{Discrimination-Skill Metric and Statistics}\label{sec:metric}
We score discrimination with chance-anchored skill
\begin{equation}
\mathrm{skill} = 2\cdot\mathrm{AUC} - 1 ,
\end{equation}
so that $0$ is chance, $1$ is perfect, and negatives are below chance. Evaluating symmetry
scores by their discrimination (AUC) follows the medical precedent of Hogeweg et~al.~\cite{hogewegFastEffectiveQuantification2017}; the rescaling $2\cdot\mathrm{AUC}-1$ is the
classical Somers'~$D$ (Gini) transform, which simply moves the chance point from $0.5$ to $0$,
so that skills are directly readable as distance from guessing. The AUC is computed
within images and then macro-averaged: for each image we take the fraction of that
image's wrong axes that the true axis outranks (ties count $\tfrac{1}{2}$), and we then average
these per-image values so that every image counts equally, whether it contributed few or many
usable negatives. Because the statistic depends only on the rank order of scores within an
image, the 13 methods' arbitrary, method-specific score scales require no
calibration. The raw score scales differ per method: for the crop of Figure~\ref{fig:task},
the three leading methods return raw scores of $0.52$, $0.66$, and $0.94$, respectively, yet
each ranks that axis perfectly against all of its negatives. All 13 scorers are constructed so that higher means more symmetric; rather than
trusting this, the harness infers each method's score direction once from its pooled scores
and holds it fixed for every analysis slice. Because the direction is inferred from the
evaluation data, a method at exactly chance level takes whichever direction looks better,
inflating near-chance skills by about $0.8$ standard errors, negligible for the leaders but
worth noting for the weakest methods.

Two aggregate statistics appear in this paper, and we label them consistently. The
\emph{leaderboard} reports the unweighted mean of the per-dataset skills. \emph{Significance}
between two methods uses a \emph{paired bootstrap}: both methods are evaluated on the same
images; we resample those images with replacement many times, recompute the skill difference
on each resample, and read a 95\% confidence interval from the spread of the resampled
differences: an interval excluding zero means the observed difference is unlikely to be
sampling luck. Because the pairing compares the two methods on identical images, between-image
difficulty cancels out, making this test considerably more sensitive than comparing two
independent confidence intervals. Under the per-axis (multi-axis)
protocol, the bootstrap unit is the axis, so datasets with several axes per image (DENDI) have
confidence intervals that are slightly too narrow; our multi-axis conclusions are
interval-overlap claims, which wider intervals would only strengthen. A cluster bootstrap that resamples whole
images (with all their axes) confirms this: per-dataset intervals widen by factors of
$1.0$--$1.4$, reinforcing the overlap-based conclusions.

\section{The Contenders: Thirteen Scoring Methods}\label{sec:scoremethods}
We evaluate 13 scoring methods (11 hand-crafted and 2 reading out learned features) spanning
the classical-to-deep progression of visual representations. Although developed independently
across three decades and several fields, all 13 instantiate a single template,
\begin{equation}\label{eq:template}
s(I) \;=\; \mathrm{A}\bigl(\mathrm{C}\bigl(T(I),\,T(M\,I)\bigr)\bigr),
\end{equation}
where $M$ is the (fixed) left--right mirror, $T$ maps the image into a representation,
$\mathrm{C}$ compares the representation of the image with that of its mirror, and $\mathrm{A}$
aggregates the comparison into one scalar. The methods differ essentially in the
\emph{representation} $T$, so we organize them into seven families, ordered from raw pixels to
learned features (Table~\ref{tab:methods}); the hand-crafted/learned split falls out as
families one through six versus seven. Each description below states the exact configuration
used throughout the benchmark; we refer to each method by the short name in
Table~\ref{tab:methods}, with its identifier in our library \texttt{imgsym} given in parentheses at
each first mention. The pool deliberately mixes provenance: six methods are faithful
(re)implementations of published measures; three are \emph{adapted variants}, each modifying
a single published measure in a disclosed way; and four have \emph{no single published counterpart}, being assembled here
from published components or generalized beyond their source's scope.
Table~\ref{tab:methods} marks
each, and every deviation from a source is stated below.
Deviations forced only to run a method on our inputs do not count as adaptation.
Exactly two methods are \emph{tuned on benchmark data}: HOG's
descriptor grid and DeepFeat's backbone and stage were selected on the single-axis ablation subsets of
Section~\ref{sec:principle} rather than taken from a source; every other method runs at its
source's stated or recommended configuration; where the source reports a parameter study,
this means its reported optimum (e.g., %
AlexNet-C2's layer and grid follow Brachmann and
Redies' stated recommendation; PatchNN's scale pair Hogeweg et~al.'s grid optimum), not an
arbitrary default. The asymmetry this leaves (two methods tuned on benchmark data, the rest at
source-recommended settings) is checked directly %
under Limitations in Section~\ref{sec:discussion}:
on data held out from both tuning pools, the leading methods keep their standing. The sources' native inputs are
diverse (gray-level brain volumes, chest radiographs, face photographs, binary abstract
patterns, paintings, and album covers), so each description names the native domain where it
differs from our natural photographs and artworks; all methods receive the same canonical
crops and convert to their working representation (typically grayscale) internally.

\begin{table}[H]
\caption{The 13 symmetry-scoring methods by representation family, ordered from raw pixels to
learned features. All instantiate Equation~\eqref{eq:template}: representation $T$, comparison
$\mathrm{C}$
, aggregation $\mathrm{A}$, and the benchmark setting of
each method's main free parameters (Configuration). 
The first 11 methods are hand-crafted; the last 2 read out frozen ImageNet-pretrained features. 
\label{tab:methods}}
\setlength{\cellWidtha}{\fulllength/5-2\tabcolsep-0in}
\setlength{\cellWidthb}{\fulllength/5-2\tabcolsep-0in}
\setlength{\cellWidthc}{\fulllength/5-2\tabcolsep-0in}
\setlength{\cellWidthd}{\fulllength/5-2\tabcolsep-0in}
\setlength{\cellWidthe}{\fulllength/5-2\tabcolsep-0in}
 \begin{adjustwidth}{-\extralength}{0cm}
  \begin{tabularx}{\fulllength}{>{\raggedright\arraybackslash}m{\cellWidtha}>{\raggedright\arraybackslash}m{\cellWidthb}>{\raggedright\arraybackslash}m{\cellWidthc}>{\raggedright\arraybackslash}m{\cellWidthd}>{\raggedright\arraybackslash}m{\cellWidthe}}
   \toprule

\textbf{Method} & \textbf{Representation \boldmath$T$} & \textbf{Comparison \boldmath$\mathrm{C}$} & \textbf{Aggregation \boldmath$\mathrm{A}$} & \textbf{Configuration}\\
\midrule
\multicolumn{5}{@{}l}{\emph{Intensity}}\\ %
PixCorr\,$^{\dag}$~\cite{mayerQuantifyingVisualAesthetics2018} & grayscale image & global cosine with mirror & ---(global) & whole image\\
SlideWin\,$^{\star}$ (after~\cite{guerriniImageSymmetriesRight2017,gnuttiCombiningAppearanceGradient2021}) & reflected half columns & Pearson correlation & overlap-weighted mean & center axis\\
EROS~\cite{smithAccurateRobustSymmetry1999} & row intensity profiles & even/odd energy (\emph{implicit}) & mean over rows & contrast-corrected\\\midrule
\multicolumn{5}{@{}l}{\emph{Binary shape}}\\
WBS~\cite{bauerlyComputationalModelingExperimental2006,gartusPredictingPerceivedVisual2017} & Otsu foreground mask & mirrored-pixel agreement & axis-weighted count ratio & vertical axis\\\midrule
\multicolumn{5}{@{}l}{\emph{Gradient field}}\\
Grad\,$^{\dag}$~\cite{gnuttiCombiningAppearanceGradient2021} & Sobel magnitude + orientation & specularity-weighted correlation & mean over rows & $90^{\circ}$ window\\
MS-Grad\,$^{\star}$ (after~\cite{gnuttiCombiningAppearanceGradient2021}) & two-scale Sobel fields & mirror-aware cosine + magnitude difference & edge-weighted map mean & 3~${\times}$~3 + 5~${\times}$~5\\\midrule
\multicolumn{5}{@{}l}{\emph{Frequency domain}}\\
DCT~\cite{gunluSymmetryAnalysis2D2009,kiryati1998detecting} & 2-D DCT coefficients & even-coefficient parity (\emph{implicit}) & energy ratio & AC terms only\\\midrule
\multicolumn{5}{@{}l}{\emph{Oriented filter banks and histograms}}\\
Gabor~\cite{shakerNewSymmetryMeasure2015} & Gabor-bank block energies & orientation energies $(\alpha,180^{\circ}{-}\alpha)$ (\emph{paired}) & mean over blocks/pairs & 12 orientations\\
HOG\,$^{\dag}$~\cite{renero-cExtractingSymmetryHuman2017,dalalHistogramsOrientedGradients2005} & dense HOG descriptor & descriptor cosine with mirror & ---(whole descriptor) & cell 16, 18 bins, unsigned\\
PHOG\,$^{\star}$ (after~\cite{renero-cExtractingSymmetryHuman2017,boschRepresentingShapeSpatial2007}) & pyramidal edge-orientation histograms & descriptor cosine with mirror & ---(whole descriptor) & 18 bins, 4 levels\\\midrule
\multicolumn{5}{@{}l}{\emph{Patch correspondence}}\\
PatchNN~\cite{hogewegFastEffectiveQuantification2017} & patch + position descriptors & bidirectional nearest-neighbor distance & mean over patches, $1/(1{+}c)$ & patch 15, $w$ 17.7, $\kappa{=}16$\\
\midrule
\multicolumn{5}{@{}l}{\emph{Learned features}}\\
AlexNet-C2~\cite{brachmannUsingConvolutionalNeural2016} & AlexNet conv2 max-pooled maps & normalized $L_1$ agreement & mean over grid cells & 11~${\times}$~11 grid\\
DeepFeat\,$^{\star}$ (after~\cite{brachmannUsingConvolutionalNeural2016}) & frozen backbone stage-1 feature map & normalized $L_1$ agreement & ---(global ratio) & MambaOut-Base~\cite{yuMambaOutWeReally}, 224 px\\
   \bottomrule
  \end{tabularx}
 \end{adjustwidth}

\noindent\footnotesize{$^{\star}$~no single published counterpart: assembled or generalized here,
after the cited sources; $^{\dag}$~adapted variant of the cited source; unmarked: faithful
(re)implementation}

\end{table}

The comparison $\mathrm{C}$ takes one of three concrete forms: an \emph{explicit mirror}
($T(I)$ compared with $T(M\,I)$ directly), an \emph{even/odd decomposition} (the even and odd
parts of a signal about the axis are the symmetric and antisymmetric components of the pair,
so parity energies compare the two implicitly), or an \emph{orientation pairing} (a mirror
maps orientation $\alpha$ to $180^{\circ}{-}\alpha$, so oriented responses are compared across
paired orientations). The three are equivalent views of one pairing rather than different
quantities: a half-image comparison, as some sources formulate it, is the explicit mirror
restricted to one side, and a parity-energy ratio equals whole-image agreement up to
normalization. Whatever its internal form, every method quantifies the $I$-versus-$M\,I$
agreement of Figure~\ref{fig:task}.

\subsection{Intensity}
\textbf{PixCorr} (\texttt{pixel\_correlation}) computes the global uncentered (cosine) correlation between the %
grayscale image and its mirror, our whole-image variant of the half-image Pearson reflection
correlation used as an aesthetics predictor by Mayer and
Landwehr~\cite{mayerQuantifyingVisualAesthetics2018}.
\textbf{SlideWin} (\texttt{sliding\_window}) compares the two reflected halves at the candidate axis with a
configurable similarity (Pearson correlation by default), weighted by overlap width; an
in-house harness with a pluggable metric that formalizes the generic scan-and-score
construction underlying axis analyses such as Guerrini et~al.~\cite{guerriniImageSymmetriesRight2017,gnuttiCombiningAppearanceGradient2021}.
\textbf{EROS} (\texttt{eros}) implements the measure of Smith and
Jenkinson~\cite{smithAccurateRobustSymmetry1999}, proposed for mid-sagittal-plane
estimation in brain images: per-row even/odd intensity-profile energies about the axis,
corrected for local and global contrast (specified only qualitatively in the source; we
instantiate them as the row and global intensity variance) and summed over rows.

\subsection{Binary Shape}
\textbf{WBS} (\texttt{weighted\_binary}) implements the perceptual measure originated by
Bauerly and Liu~\cite{bauerlyComputationalModelingExperimental2006}, specified as implemented
here by Gartus and Leder~\cite{gartusPredictingPerceivedVisual2017} and also employed by
H\"ubner and Fillinger~\cite{hubnerComparisonObjectiveMeasures2016}: matching mirrored
foreground pixels are counted with weights increasing linearly toward the axis. Its native
inputs are abstract black-and-white patterns, where the foreground is the image;
natural images first need a foreground mask, which we obtain by Otsu thresholding, the
simplest instantiation, chosen so the measure itself stays faithful (a learned
foreground segmenter could be substituted at the cost of a heavier pipeline). The benchmark
evaluates its vertical-axis component for a like-for-like left--right~comparison.

\subsection{Gradient Field}
\textbf{Grad} (\texttt{gradient}) weights Sobel magnitudes by an orientation-specularity factor (the gradient
directions of a mirrored pixel pair must be specular about the axis) and correlates the
weighted map with its mirror row by row, our simplified, global variant of the
appearance-plus-gradient criterion of Gnutti et~al.~\cite{gnuttiCombiningAppearanceGradient2021}, whose original is a candidate-axis
selection-and-validation pipeline. \textbf{MS-Grad} (\texttt{multi\_scale\_gradient}; in-house,
after~\cite{gnuttiCombiningAppearanceGradient2021}) fuses two Sobel scales with
texture-adaptive weights and scores a mirror-aware cosine agreement (under $M$ the
$x$-gradient flips sign while the $y$-gradient is preserved) plus a normalized magnitude
difference, pooled through an edge-weighted loss map.

\subsection{Frequency Domain}
\textbf{DCT} (\texttt{dct}) measures the fraction of AC energy held by the even-indexed DCT columns, the
parity signature of reflection symmetry, following Gunlu and
Bilge~\cite{gunluSymmetryAnalysis2D2009}, who proposed it for locating symmetry axes in face
and texture images; it realizes, in the transform domain, the
symmetric/antisymmetric energy decomposition that Kiryati and Gofman formulated in the spatial
domain~\cite{kiryati1998detecting}.

\subsection{Oriented Filter Banks and Histograms}
\textbf{Gabor} (\texttt{gabor}), proposed for the aesthetic evaluation of paintings, applies a
12-orientation Gabor bank (the source specifies the
kernels only symbolically; our parameterization is disclosed in the released library),
thresholds the magnitude responses into binary partial images at the source's threshold, and
compares block energies between each orientation $\alpha$ and the mirrored partial image of the
paired orientation $180^{\circ}-\alpha$~\cite{shakerNewSymmetryMeasure2015}. \textbf{HOG} (\texttt{hog}) adapts the
face-symmetry measures of Renero-C et~al.~\cite{renero-cExtractingSymmetryHuman2017} (there,
averages of inner products of per-region HOG descriptors~\cite{dalalHistogramsOrientedGradients2005}, over landmark patches or, in their
best-performing variant, whole-face vertical strips) into one pooled, dense whole-image
descriptor compared with its mirror by a single
cosine; this formulation, and its tuned configuration (cell 16; 18 bins; unsigned gradients)
found by the ablation of Section~\ref{sec:principle}, are ours.
\textbf{PHOG} (\texttt{phog}) is our combination of the same mirror-cosine measure~\cite{renero-cExtractingSymmetryHuman2017} with the
pyramidal, Canny-gated orientation histogram of Bosch et~al.~\cite{boschRepresentingShapeSpatial2007}.

\subsection{Patch Correspondence}
\textbf{PatchNN} (\texttt{local\_global}) samples mirrored patch pairs about the axis, forms z-scored
patch-plus-position descriptors with a folded-$x$ coordinate, and maps the mean bidirectional
nearest-neighbor distance between the left and right descriptor sets to a score, the
chest-radiography symmetry quantification of Hogeweg et~al.~\cite{hogewegFastEffectiveQuantification2017}, whose nearest-neighbor matching deliberately
does not assume pixel-exact symmetry. Our implementation runs the source's grid optimum
(patch size $15$; position weight $17.7$) at its $\kappa~{=}~16$ sampling stride, substituting
an exact nearest-neighbor search for the original's approximate kd-tree (the source's
full-sampling default is intractable under exact search).

\subsection{Learned Features}
\textbf{AlexNet-C2} (\texttt{alexnet}) faithfully reimplements Brachmann and
Redies~\cite{brachmannUsingConvolutionalNeural2016} at their stated general recommendation:
they evaluate layers conv1--conv5 and advise second-layer features for arbitrary images, so
second-layer filter responses of a pretrained AlexNet are adaptive-max-pooled to their
reported conv2 optimum, an 11~${\times}$~11 patch grid, for the image and its mirror, and
compared by normalized $L_1$ agreement (their \mbox{Equations~(1) and (2)}). \textbf{DeepFeat} (\texttt{deep\_features}) is our generalization of the same construction from one fixed layer to
arbitrary frozen \texttt{timm}~\cite{rw2019timm} backbones and stages; the benchmark default is a MambaOut-Base
stage-1 feature map~\cite{yuMambaOutWeReally}. Neither
network is fine-tuned: both scorers read symmetry out of frozen ImageNet-pretrained
representations.

\section{Results}\label{sec:results}
We organize the results as four questions: which representations discriminate mirror symmetry
best (Section~\ref{sec:leaderboard}); whether the ranking is an artifact of the protocol
(Section~\ref{sec:robustness}); why the winners win (Section~\ref{sec:principle}); and what
scoring costs in practice (Section~\ref{sec:practice}).

\subsection{Which Representations Discriminate Mirror Symmetry Best?}\label{sec:leaderboard}
\begin{figure}[H]
\includegraphics[width=11.5cm]{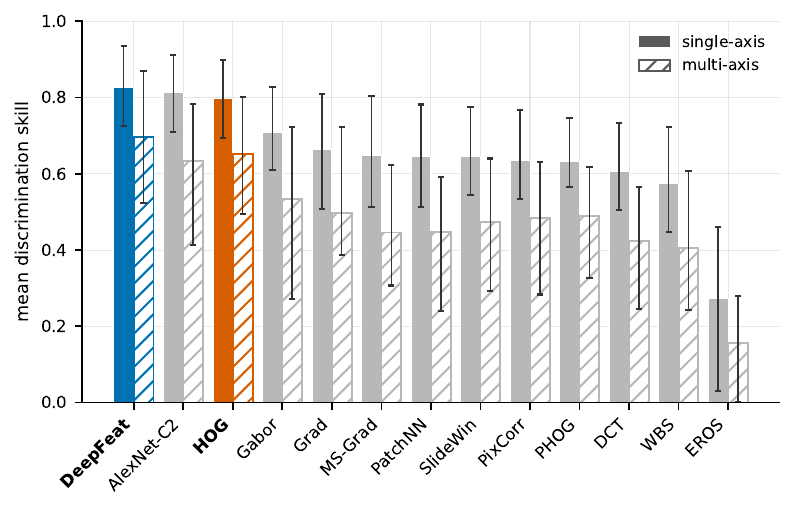}
\caption{Mean discrimination skill of all 13 methods under the single-axis (solid) and
multi-axis (hatched) protocols; whiskers span the per-dataset range. DeepFeat
(blue) and HOG (orange), the best learned and the best hand-crafted method, are shown in colour
with bold labels; all other methods are gray.\label{fig:svm}}
\end{figure}

The leaderboard (Table~\ref{tab:leaderboard}, Figure~\ref{fig:svm}) puts learned features on
top, with the CNN-filter measure and the tuned oriented-gradient histogram not statistically
separable behind them, and everything else well behind. DeepFeat leads at a mean single-axis skill
of $0.83$; AlexNet-C2 ($0.81$) and HOG ($0.80$) follow, not statistically separable from each other; a clear
gap follows to Gabor ($0.71$) and Grad ($0.67$), a broad mid-pack drawn from
every hand-crafted family sits at $0.58$--$0.65$, and EROS (a brain-midplane measure operating far
outside its native domain) marks the floor at $0.27$. The paired bootstrap, with
Holm correction over the six-comparison ladder (the three pairwise comparisons
among the top three methods, plus HOG versus Gabor, Gabor versus Grad, and Grad versus PatchNN), sharpens the top: DeepFeat's lead over HOG is
small but significant and survives correction ($+0.031$, Holm-adjusted $p=0.044$); its nominal
lead over AlexNet-C2 ($+0.016$, unadjusted $p=0.041$, Holm-adjusted $p=0.123$) does not; and AlexNet-C2 and HOG are not
significantly different ($+0.015$), although formal equivalence within $\pm0.03$ is not
established either (TOST $90\%$ CI $[-0.005,+0.035]$): the three
leaders are separated only by DeepFeat's edge over HOG. Below the leaders, HOG beats Gabor
decisively ($+0.092$), and Gabor $>$ Grad ($+0.041$) is the last significant step; the mid-pack
is not separable.

\begin{table}[H]
\caption{Mean discrimination skill ($2\cdot\mathrm{AUC}-1$; four single-axis and five
multi-axis datasets) under both protocols, sorted by single-axis skill, with 95\% bootstrap
confidence intervals on each mean (units resampled within datasets, direction fixed per
method). 
Bold marks the best learned method (DeepFeat) and the best hand-crafted method (HOG). Per-dataset values are given
in Appendix~\ref{app:perdataset}.%
\label{tab:leaderboard}}
\begin{tabularx}{\textwidth}{lCC}
\toprule
\textbf{Method} & \textbf{Single-Axis} & \textbf{Multi-Axis}\\
\midrule
\textbf{{DeepFeat}} & +0.83 [+0.81, +0.85] & +0.70 [+0.68, +0.71]\\ 
AlexNet-C2 & +0.81 [+0.79, +0.83] & +0.63 [+0.61, +0.66]\\
\textbf{{HOG}} & +0.80 [+0.77, +0.82] & +0.65 [+0.63, +0.67]\\
Gabor & +0.71 [+0.68, +0.74] & +0.53 [+0.50, +0.57]\\
Grad & +0.67 [+0.63, +0.70] & +0.50 [+0.46, +0.53]\\
MS-Grad & +0.65 [+0.61, +0.68] & +0.45 [+0.41, +0.48]\\
PatchNN & +0.65 [+0.61, +0.68] & +0.45 [+0.41, +0.48]\\
SlideWin & +0.65 [+0.61, +0.68] & +0.47 [+0.43, +0.51]\\
PixCorr & +0.64 [+0.60, +0.67] & +0.48 [+0.45, +0.51]\\
PHOG & +0.63 [+0.60, +0.66] & +0.49 [+0.46, +0.52]\\
DCT & +0.61 [+0.57, +0.64] & +0.42 [+0.39, +0.46]\\
WBS & +0.58 [+0.54, +0.61] & +0.40 [+0.37, +0.44]\\
EROS & +0.27 [+0.23, +0.32] & +0.16 [+0.12, +0.19]\\
\bottomrule
\end{tabularx}

\noindent\footnotesize{The intervals are per-method, so overlap between two methods does not contradict a significant \emph{paired} difference; method-versus-method conclusions rest on the paired bootstrap reported in the text.}
\end{table}
\vspace{-12pt}

\subsubsection*{Sensitivity to Perturbation Magnitude}
\begin{figure}[H]
\includegraphics[width=\textwidth]{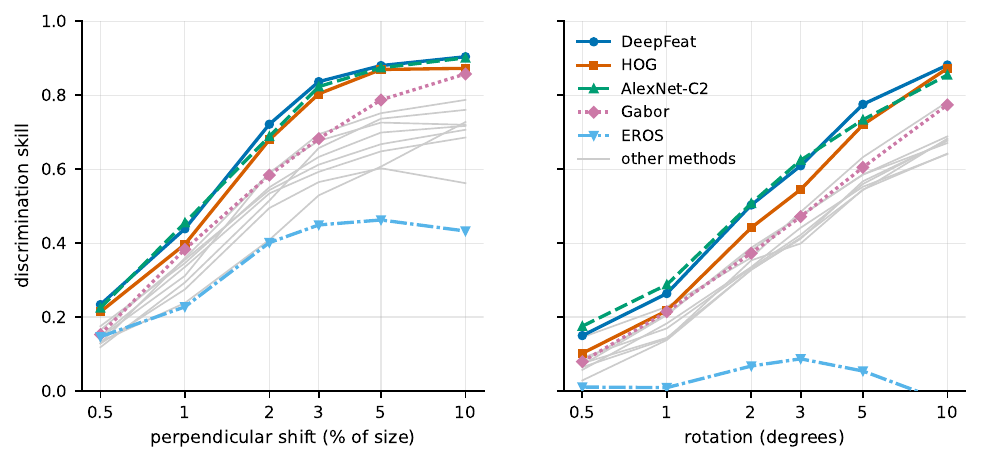}
\caption{Discrimination skill vs.\ perturbation magnitude (\textbf{left}: perpendicular shift, \% of
size; \textbf{right}: rotation, degrees). All 13 methods are shown (gray = context); the three leaders,
Gabor (mid-pack), and EROS (floor) are highlighted.\label{fig:sensitivity}}
\end{figure}

Sweeping the perturbation magnitude shows where the ranking is decided
(Figure~\ref{fig:sensitivity}). All methods sit near chance at the finest perturbations (skills
of $0.12$--$0.24$ at a $0.5\%$ shift, at most $0.18$ at $0.5^{\circ}$) and near their ceilings
at the coarsest ($10\%$/$10^{\circ}$), so the methods separate most in the
$1$--$3\%$/$1$--$3^{\circ}$ band. DeepFeat rises fastest (at a $2\%$ shift it already reaches
$0.73$ versus HOG's $0.67$), making it the most precise localizer of the true axis. Rotation is
almost uniformly harder than translation: every method except
PHOG (and WBS and HOG at the coarsest magnitude) scores lower against rotated negatives at
matched nominal levels (DeepFeat: $0.52$ at $2^{\circ}$ vs.\ $0.73$ at $2\%$), and EROS fails
rotated negatives almost completely (skill ${\leq}0.10$ at every angle), a structural
blindness, not an input mismatch: its independent per-row profile score is nearly invariant to
rotations about the crop center, an error mode its native use (searching mid-plane
orientations in near-globally symmetric brain volumes) never exercises.

\subsection{Is the Ranking an Artifact of the Protocol?}\label{sec:robustness}
Four checks say no. Fixed-extent crops leave the ranking essentially unchanged
(Section~\ref{sec:extraction}); the per-axis protocol below preserves it on five further
datasets; stratifying the negatives shows the leading methods behave as designed while
exposing the failure mode of the weakest; and corrupting the input images themselves leaves
the leader tier intact.

\subsubsection{Generalization to the Multi-Axis Protocol}\label{sec:multiaxis}
The leaderboard above scores one dominant axis per image. We repeat the benchmark under the
\emph{per-axis} protocol of Section~\ref{sec:datasets} (scoring every annotated reflection
axis, and adding the DENDI
dataset~\cite{seoReflectionRotationSymmetry2022}) to test whether the finding is an artifact of
the single-axis setup. It is not (Figure~\ref{fig:svm}; Table~\ref{tab:leaderboard}, multi-axis
column): the method ranking is preserved. DeepFeat stays first, with HOG and AlexNet-C2 as the
next two (swapping order, consistent with their non-separation), and DeepFeat and HOG retain
overlapping 95\% confidence intervals on all five datasets (Appendix~\ref{app:perdataset});
the same drop to the mid-pack
follows. Multi-axis discrimination is uniformly harder (every skill drops, and DENDI's best
methods reach only ${\sim}0.5$), so the benchmark is far from saturated. On DENDI, the dataset
introduced by the state-of-the-art learned detector, the tuned classical HOG ($0.49$)
stays within the same small margin of the deep method ($0.52$) as on the single-axis sets
(paired difference $+0.027$, $90\%$ CI $[+0.009,+0.045]$): the classical descriptor remains
competitive on the learned method's own data.

\subsubsection{Local Versus Global Discrimination}\label{sec:localglobal}
Stratifying the negatives separates two abilities: \emph{local} skill (rank the true axis above
its small perturbations, hard) and \emph{global} skill (reject gross structural alternatives,
easy). With near-coincident negatives excluded (Section~\ref{sec:perturb}), global skill exceeds
local skill for every competent method, as it should (DeepFeat $+0.61$ local/$+0.88$
global; HOG $+0.56$/$+0.88$; AlexNet-C2 $+0.59$/$+0.91$, the strongest global rejection of any
method). Three methods inverting the pattern fail
exactly this sanity check: PatchNN ($+0.44$/$+0.29$), WBS ($+0.41$/$+0.38$),
and EROS ($+0.19$/$+0.09$) localize adequately yet cannot reject gross alternatives. PatchNN and
WBS, like EROS, run outside their native domains (chest radiographs; binary patterns), so these
failures conflate method weakness with domain mismatch.

\subsubsection{Robustness to Input Degradation}\label{sec:degradation}
Finally, we corrupt the input images themselves. Each full image is degraded before crop
extraction (the pipeline downstream is unchanged) with four degradation families at three
levels each: Gaussian blur ($\sigma\in\{1,2,4\}$~px), additive Gaussian noise
($\sigma\in\{5,10,20\}$ on the 0--255 scale), JPEG re-compression (quality
$\in\{50,25,10\}$), and mean-fill square occlusion (side $\in\{10,20,30\}\%$ of the short
image side, position seeded per image). All 13 methods are then re-run on the four
single-axis datasets under the standard negatives and aggregation of
Table~\ref{tab:leaderboard} (the single-axis protocol; Section~\ref{sec:multiaxis} shows the
ranking is protocol-invariant). Table~\ref{tab:degradation} reports mean skill per
condition; combined 95\% confidence half-widths on these means are $0.02$--$0.04$, with
per-dataset values and bootstrap intervals in the released results.

Four regularities stand out. First, the leader tier is robust: the top three remain the top
three under every condition, and DeepFeat in particular is indifferent to blur, noise, and
compression ($0.82$--$0.84$ throughout, against $0.83$ clean). Second, noise separates the
representations: it is HOG's one weakness ($0.80\to0.71$ at $\sigma=20$; pixel noise corrupts
the fine gradient histogram) while the network readouts pool it away (AlexNet-C2 $-0.03$;
DeepFeat $+0.01$), widening DeepFeat's margin over HOG from $0.03$ to $0.13$: on noisy
inputs, the deep premium grows. Third, blur fells the fixed-scale edge and filter methods:
PHOG collapses to near chance ($0.63\to0.08$ at $\sigma=4$; its Canny edge pyramid vanishes)
and Gabor loses a tier ($0.71\to0.49$; a fixed-wavelength bank loses its band), while the
other eleven methods lose between $0.01$ (DeepFeat) and $0.12$ (EROS), the leaders at most
$0.07$ (AlexNet-C2). JPEG
re-compression, in contrast, is harmless everywhere: no method loses more than $0.02$, and
HOG even gains. Fourth, occlusion is the corruption that reaches everyone (at $30\%$ all 13
methods lose skill), and it converges the leaders: DeepFeat's edge over
AlexNet-C2 dissolves ($0.74$ both), with HOG at $0.71$ retaining its usual small deficit. %
For
deployment, prefer a network readout when inputs are noisy, avoid edge- and filter-bank
methods under blur, and expect no frozen-deep premium over AlexNet-C2 \mbox{($10$ vs.\
${\sim}170$~ms per crop)} when inputs are heavily occluded.

\begin{table}[H]
\caption{Discrimination skill under input degradation: mean over the four single-axis
datasets (the protocol and aggregation of Table~\ref{tab:leaderboard}), for all 13 methods
under four degradation families at three levels each (clean = uncorrupted baseline).
Combined 95\% confidence half-widths on the means are $0.02$--$0.04$; per-dataset values with
bootstrap intervals are in the released results. Bold as in Table~\ref{tab:leaderboard}.\label{tab:degradation}} %
\begin{adjustwidth}{-\extralength}{0cm}
\footnotesize
\begin{tabularx}{\fulllength}{lCCCCCCCCCCCCC}
\toprule
 \multirow{2}{*}{\textbf{Method}} &  \multirow{2}{*}{\textbf{Clean}} & \multicolumn{3}{c}{\textbf{Blur} \boldmath$\sigma$} & \multicolumn{3}{c}{\textbf{Noise} \boldmath$\sigma$} & \multicolumn{3}{c}{\textbf{JPEG} \boldmath$q$} & \multicolumn{3}{c}{\textbf{Occlusion \%}}\\
 & & \textbf{1} & \textbf{2} & \textbf{4} & \textbf{5} & \textbf{10} & \textbf{20} & \textbf{50} & \textbf{25} & \textbf{10} & \textbf{10} & \textbf{20} & \textbf{30}\\
\midrule
\textbf{{DeepFeat}} & +0.83 & +0.83 & +0.84 & +0.82 & +0.83 & +0.83 & +0.84 & +0.84 & +0.84 & +0.84 & +0.82 & +0.79 & +0.74\\ 
AlexNet-C2 & +0.81 & +0.80 & +0.78 & +0.74 & +0.81 & +0.80 & +0.79 & +0.81 & +0.81 & +0.80 & +0.81 & +0.79 & +0.74\\
\textbf{{HOG}} & +0.80 & +0.81 & +0.81 & +0.77 & +0.77 & +0.75 & +0.71 & +0.80 & +0.81 & +0.82 & +0.79 & +0.76 & +0.71\\
Gabor & +0.71 & +0.71 & +0.63 & +0.49 & +0.71 & +0.70 & +0.67 & +0.70 & +0.71 & +0.69 & +0.71 & +0.69 & +0.65\\
Grad & +0.67 & +0.67 & +0.66 & +0.59 & +0.67 & +0.66 & +0.63 & +0.67 & +0.67 & +0.69 & +0.67 & +0.64 & +0.61\\
MS-Grad & +0.65 & +0.68 & +0.66 & +0.60 & +0.64 & +0.62 & +0.57 & +0.65 & +0.65 & +0.65 & +0.64 & +0.61 & +0.55\\
PatchNN & +0.65 & +0.66 & +0.65 & +0.63 & +0.64 & +0.63 & +0.62 & +0.65 & +0.65 & +0.66 & +0.65 & +0.62 & +0.58\\
SlideWin & +0.65 & +0.64 & +0.63 & +0.60 & +0.64 & +0.65 & +0.64 & +0.64 & +0.64 & +0.64 & +0.64 & +0.63 & +0.61\\
PixCorr & +0.64 & +0.64 & +0.63 & +0.61 & +0.64 & +0.64 & +0.64 & +0.64 & +0.64 & +0.64 & +0.64 & +0.63 & +0.61\\
PHOG & +0.63 & +0.60 & +0.46 & +0.08 & +0.63 & +0.64 & +0.62 & +0.62 & +0.62 & +0.62 & +0.62 & +0.60 & +0.59\\
DCT & +0.61 & +0.60 & +0.59 & +0.57 & +0.60 & +0.60 & +0.60 & +0.60 & +0.61 & +0.60 & +0.60 & +0.59 & +0.57\\
WBS & +0.58 & +0.58 & +0.56 & +0.55 & +0.57 & +0.57 & +0.57 & +0.58 & +0.56 & +0.57 & +0.57 & +0.55 & +0.50\\
EROS & +0.27 & +0.25 & +0.21 & +0.15 & +0.27 & +0.27 & +0.26 & +0.27 & +0.27 & +0.27 & +0.27 & +0.24 & +0.21\\
\bottomrule
\end{tabularx}

\end{adjustwidth}
\end{table}

\subsection{Why Do the Winners Win? The Mid-Scale Oriented-Feature Principle}\label{sec:principle}
\begin{figure}[H]
\includegraphics[width=\textwidth]{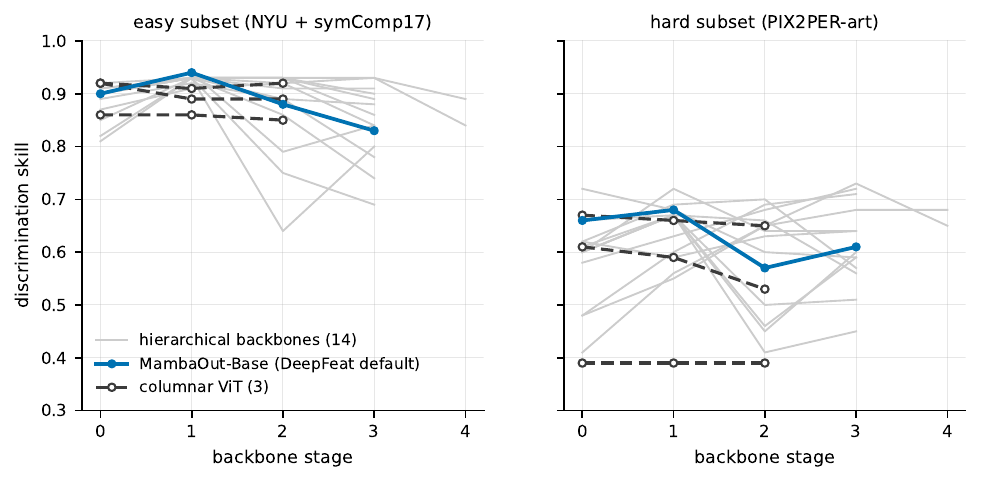}
\caption{Discrimination skill by backbone stage across 17 backbones on the easy (NYU + symComp17, \textbf{left}) and
hard (PIX2PER-art, \textbf{right}) subsets; each line ends at its backbone's last stage (backbones expose three to
five stages). Hierarchical networks are gray, MambaOut-Base (DeepFeat's default backbone) is
highlighted in blue, and the three columnar ViTs (ViT-Tiny/16, ViT-Base/16, and the
finer-patch control ViT-Base/8) are dashed with open markers. All 17 backbones are listed in Appendix~\ref{app:backbones}.\label{fig:stage}}
\end{figure}

Two independent ablations locate the signal at mid scales. First, we sweep the stage at
which DeepFeat reads its feature map, across every stage of 17 pretrained backbones spanning diverse modern architectures (the full
list with per-subset optima is given in Appendix~\ref{app:backbones}). On an easy subset (NYU + symComp17; 40 images each; $n=80$), the 14 hierarchical backbones peak at a low or mid stage and decline toward their deepest, most semantic stage; ConvNeXtV2-Base and MambaOut-Base lead at $0.94$, %
with eleven more backbones clustered at $0.93$ within overlapping intervals: it is the layer,
not the architecture, that matters (Figure~\ref{fig:stage}). The columnar Vision
Transformers (ViT-Tiny/16 and ViT-Base/16, joined by a finer-patch ViT-Base/8 control from the
same AugReg weights family), which process tokens at a single spatial resolution and lack a
spatial hierarchy, never develop a mid-stage peak, and the patch-16 pair closes the ranking.
Two readings are open at this point: the missing hierarchy, or the coarse $16$-px tokenization
discarding mid-scale structure at the embedding. A harder subset
(PIX2PER-art, $n=40$) confirms the picture while adding a caveat: the hierarchical backbones
cluster at $0.64$--$0.73$ with overlapping intervals (EfficientNetV2-T at $0.73$, then
ConvNeXtV2-Atto, EfficientViT-B3, and MambaOut-Tiny at $0.72$; no single backbone dominates), and
the patch-16 ViTs are now clearly worst ($0.39$ and $0.61$), but the exact optimal stage becomes
architecture-dependent (stages~0--3, rarely the deepest). The ViT-Base/8 control separates the
two readings: with $8$-px tokens and no other change, the columnar model recovers to $0.67$ on
the hard subset, inside the hierarchical cluster, while staying flat across stages
($0.67/0.66/0.65$; on the easy subset it matches ViT-Base/16 at $0.92$). At $n=40$ the recovery
is not individually significant ($0.67$ $[0.55,0.79]$ vs.\ ViT-Base/16's $0.61$
$[0.46,0.74]$), but its direction resolves the confound the patch-16 pair leaves open: much of
the columnar deficit is tokenization stride, not the absence of a hierarchy as such. The common
denominator across all $17$ backbones is \emph{access to mid-scale oriented structure}:
hierarchical models reach it through their low/mid stages, and a columnar model regains most
of it when its tokens are fine enough. Low/mid spatial features win; the
clean universal stage-1 peak is partly an easy-data effect. The pattern even reproduces inside
the benchmark's own AlexNet scorer: re-scored at the source's layer-1 optimum (conv1 on a
17~${\times}$~17 grid), it reaches $0.79$ against the benchmarked conv2's $0.81$, the rising
limb of the mid-stage peak within AlexNet itself.

\begin{figure}[H]
\includegraphics[width=\textwidth]{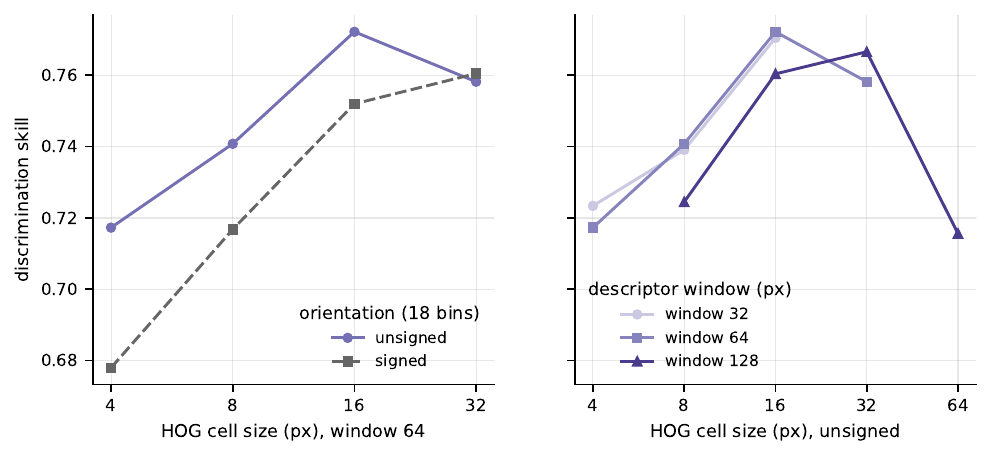}
\caption{HOG ablations. (\textbf{Left}): skill vs.\ cell size (signed and unsigned gradients) at the
default 64-px window. (\textbf{Right}): cell size swept against descriptor window size (unsigned, 18
bins).\label{fig:hogcell}}
\end{figure}

Second, the same pattern appears in a purely hand-crafted parameter: HOG's cell size. On an
easy+hard pool (symComp17 + PIX2PER-art, $n=298$) we sweep cell size $\{4,8,16,32\}$,
orientation bins $\{9,18\}$, and signed versus unsigned gradients at the default 64-px
descriptor window, plus a joint sweep of cell size against descriptor window size (windows $\{32,64,128\}$~px). Skill
rises from $0.72$ (cell 4) to a peak of $0.77$ at cell 16 and falls again at cell 32
(Figure~\ref{fig:hogcell}, left): too fine sees texture, too coarse sees only layout, and
mirror symmetry lives in between. The window sweep separates absolute from relative scale:
cell 16 remains optimal at the 32- and 64-px windows, drifts one step at 128~px, and the
best-per-window skills are not statistically separable; the peak follows (approximately)
absolute cell scale, not the cell:window ratio (Figure~\ref{fig:hogcell}, right). Unsigned
orientation beats signed at every informative resolution (by up to $+0.057$ skill, $+0.020$ at
the peak configuration) for a clean geometric reason: a mirror \emph{preserves} unsigned
orientation but \emph{flips} the signed gradient direction; at the coarsest, near-global cell, the
advantage vanishes, exactly where this argument predicts. The tuned library default (cell 16;
18 bins; unsigned) was selected on this grid; on the full four-dataset benchmark it is worth $+0.058$ pooled skill over the
descriptor's common default (significant under the paired bootstrap), and the gain holds on
datasets held out from the tuning pool.
Two knobs from two different worlds, network depth
and descriptor cell size, agree on the same mid-scale answer.
The benchmark's own PHOG makes the same point in reverse: it applies the identical
mirror-cosine comparison as HOG, but to a pyramidal descriptor that gates its histograms on
sparse Canny edges, whose finest pyramid cell is far coarser than the 16-px optimum, %
and whose orientation bins are signed. Each choice cuts against the dense, mid-scale, unsigned recipe
identified above, and PHOG trails HOG by $0.17$ skill ($0.63$ vs.\ $0.80$).

\subsection{What Does Scoring Cost in Practice?}\label{sec:practice}

\subsubsection{Speed Versus Accuracy}\label{sec:speed}
\begin{figure}[H]
\includegraphics[width=10.5cm]{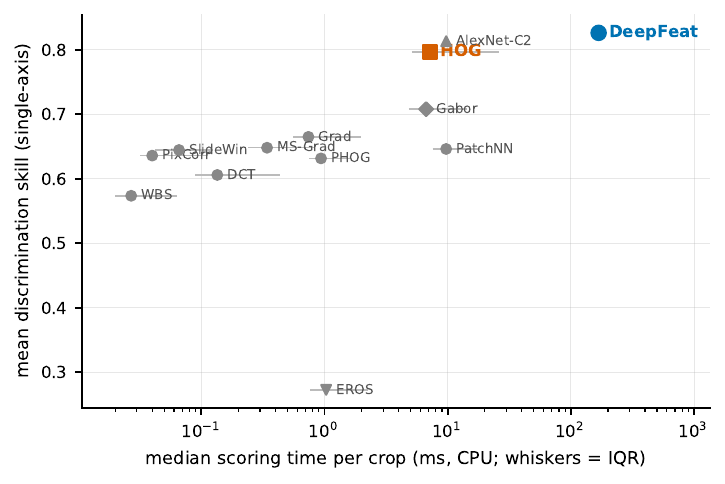}
\caption{Speed versus accuracy: median CPU scoring time per crop in milliseconds (log scale;
whiskers = interquartile range) against mean single-axis discrimination skill.\label{fig:speed}}
\end{figure}

Runtime is measured by scoring a fixed sample of 48 native-resolution true-axis crops (12 per
dataset) on a single CPU process after two warm-up calls; we report per-crop medians with
interquartile ranges. Benchmark scoring and all timings ran on one workstation, an
AMD Ryzen~7~9700X (8~cores/16~threads) with 32~GB RAM under Ubuntu~24.04 (WSL2 on
Windows~11), using Python~3.12 with PyTorch~2.12 (CPU build), \texttt{timm}~1.0.27,
OpenCV~4.13, NumPy~2.4, SciPy~1.18, and scikit-image~0.26 at library-default thread
settings. Figure~\ref{fig:speed} places every method on the cost--skill
plane, both axes describing the same operating point, since the timed crops are drawn from
the same benchmark that produces the skill numbers.
The tuned HOG scores a crop in $0.5$~ms, over $300\times$ faster than the best deep method
(${\sim}170$~ms per crop for the MambaOut stage-1 forward pass), while trailing it by only
$0.03$ skill. AlexNet-C2 edges ahead of HOG's skill ($0.81$ vs.\ $0.80$, not statistically separable)
but needs $10$~ms per crop, ${\sim}20\times$ HOG's cost: the classical descriptor delivers
accuracy not statistically separable from the cheapest network's at a fraction of its price,
and the main reasons to pay for full network inference are the final $+0.02$ it adds over
AlexNet-C2 (skill $0.83$ vs.\ $0.81$, not statistically separable) and its markedly
better blur tolerance (Section~\ref{sec:degradation}). The remaining
classical methods are beaten on both cost and skill (Gabor costs ${\sim}15$~ms and PatchNN $8$~ms for mid-pack
skill), leaving three options worth considering, one per cost tier: sub-$0.1$-ms intensity methods (skill
${\sim}0.65$), HOG at $0.5$~ms (skill $0.80$), and DeepFeat at ${\sim}170$~ms (skill $0.83$).

\subsubsection{Complementarity and Ensembles}\label{sec:ensemble}
The top methods make different mistakes: the per-image success correlation between DeepFeat and
HOG is only $0.46$, versus $0.78$ between DeepFeat and AlexNet-C2: the oriented-gradient
histogram is \emph{complementary} to the deep readout, while the two network-based scorers are
largely redundant. The ensembles agree (skills here are pooled over images, so the single-method baseline is
DeepFeat's pooled $0.819$, not its per-dataset-mean $0.83$): combining it with the redundant
AlexNet-C2 gains exactly nothing ($+0.000$), rank-combining DeepFeat with HOG raises the pool
to $0.833$, adding AlexNet-C2 on top reaches $0.840$ ($+0.021$), and pooling all 13 methods
drags the ensemble below the best single method ($0.794$).
Method rankings are moderately stable across datasets (mean pairwise Spearman correlation
$0.84$, minimum $0.73$).

\section{Discussion}\label{sec:discussion}
Do the frozen deep features of the scoring literature earn their cost? For most purposes, no.
The best frozen readout beats a tuned HOG by $0.031$ skill (real, but small) while HOG runs
${\sim}300\times$ faster on CPU and is not statistically separable from the CNN-based
AlexNet-C2 measure. We
stress the scope of this verdict: no scorer in the pool is trained for the task, so
whether a learned head on such features (or a detection network's confidence repurposed for a
given axis) can widen the gap is untested here and an open question. In practice the choice
depends on the cost one can afford (Section~\ref{sec:speed}): intensity measures
when only coarse discrimination is needed, HOG as the default choice, and a mid-stage deep
readout when the final $+0.03$ justifies two orders of magnitude more compute, or as part of an
ensemble when the best accuracy is needed (Section~\ref{sec:ensemble}). The deep option is likely cheaper than our reference figure
suggests: in the stage ablation, compact backbones (ConvNeXtV2-Atto, MambaOut-Tiny) match the
large ones within intervals on both subsets, so the ${\sim}170$-ms MambaOut-Base forward pass
is an upper bound on what mid-stage deep scoring must cost, not a requirement.

The reason appears to be representational rather than a matter of capacity. Both routes to the
mid-scale principle (Section~\ref{sec:principle}), and the ViT results point to the same
operative ingredient: \emph{access to mid-scale oriented structure}. Hierarchical models
supply it through their low and mid stages; coarse $16$-px tokenization denies it, and finer
tokens partly restore it. The unsigned-orientation result adds the geometric mechanism:
reflection preserves unsigned orientation but flips signed direction. This converges with independent perception-oriented
evidence: probing a canonical classification CNN with abstract dot patterns, Bonneh and
Tyler~\cite{bonnehCanonicalDeepNeural2025} likewise find symmetry discriminability absent in
the earliest layers, emerging at a mid layer, and insensitive to contrast polarity, although
their peak sits later (at fully connected layers) for object-free dot displays than ours does
for natural photographs. The same direction appears in Brachmann and Redies' own
data~\cite{brachmannUsingConvolutionalNeural2016}, where higher layers correlate best
with the perceived symmetry of album covers: predicting perceptual judgments rewards deeper
features, while discriminating the geometric axis rewards mid-scale ones. Orientation
histograms as carriers of mirror symmetry also have deep classical
roots~\cite{sunFastReflectionalSymmetry1999,kanezakiMirrorReflectionInvariant2014}.

One boundary is worth stating plainly. The benchmark measures relative discrimination
within an image (whether a scorer ranks the true axis above wrong ones), so it certifies whether
a method can tell \emph{which} axis is more symmetric, not \emph{how} symmetric an image is on a
scale comparable across images. The raw scores do not fill that gap: they are method-specific,
uncalibrated (Section~\ref{sec:metric}), and dominated by image content, so the same value means
different things on different images. A calibrated, cross-image degree of symmetry is a separate
task we do not take up here.

A second boundary is perceptual. The benchmark certifies geometric discrimination against
annotated axes, not agreement with human symmetry ratings, and the two are known to
dissociate: in Brachmann and Redies' data, deeper CNN layers best predict the perceived
symmetry of album covers, while our geometric task rewards mid-scale features. The
perception literature has validated pixel-level measures against human ratings of abstract
binary
patterns~\cite{bauerlyComputationalModelingExperimental2006,hubnerComparisonObjectiveMeasures2016,gartusPredictingPerceivedVisual2017}.
PIX2PER's annotations bring human perception into the benchmark's ground truth (observers
marked the axes~\cite{muradasodriozolaPixelsPerceptionBenchmark2026}), but none of the
datasets provides graded ratings of how symmetric an image looks, so rating-level
validation of the thirteen scorers remains a natural companion study.

Our claims concern scoring, not detection. On the detection side, large pretrained and
equivariant models currently
dominate~\cite{seoReflectionRotationSymmetry2022,yuAxislevelSymmetryDetection2025,yangCLIPSymDelvingSymmetry2025},
and this is not in tension with our finding that coarsely tokenized columnar-ViT features perform poorly at
scoring: detectors learn task-specific, spatially decoded readouts, whereas we probe frozen
representations with a fixed comparison. Because detection can be framed as a search over
candidate axes wrapped around a scorer (Section~\ref{sec:task}), a natural next step is to
drive that search with a strong, inexpensive scorer from this benchmark. The degradation
suite (Section~\ref{sec:degradation}) maps where the leaders are fragile: noise is the tuned
HOG's one weak spot (and widens the deep premium), blur fells the edge- and filter-bank
methods, and heavy occlusion, the one corruption that reaches all thirteen methods,
dissolves the frozen-deep edge \mbox{over~AlexNet-C2}.

\subsection*{Limitations}
\emph{Selection and configuration.} Both winning configurations (HOG's cell, bin count, and %
sign and DeepFeat's stage) were selected on single-axis benchmark subsets; the two tuned methods received
identical treatment, and the selection does not drive the conclusions: HOG's tuned
configuration generalizes (mean skill $0.81$ on the datasets outside its tuning pool, vs.\
$0.80$ pooled), and on PIX2PER-nat (the one dataset outside both tuning
pools) DeepFeat ($0.77$) and AlexNet-C2 ($0.77$) still lead HOG ($0.69$). The remaining
methods run at their source's recommended or default configuration, so exact optima should not
be~over-read.

\emph{Statistical caveats.} Two data-dependent choices are quantified where they are introduced
(Section~\ref{sec:metric}): direction inference gives near-chance methods a ${\sim}0.8$-standard-error
positive bias, irrelevant at the leaders' levels, and the per-axis bootstrap unit is checked
by an image-level cluster bootstrap that widens the multi-axis intervals only mildly, preserving
every overlap claim. One floor we cannot measure: at the finest perturbation magnitudes
(${\leq}1\%$/$1^{\circ}$), the annotated axis may no longer be the most symmetric line in the
image, so annotation noise and method precision are confounded there (one reason the main
results use the coarser half of the ladder), and PIX2PER distributes only consensus
annotations, so the inter-rater axis spread that would quantify this floor is unavailable.

\emph{Scope.} The images are natural photographs and
artworks; the ranking need not carry over to specialized domains such as grayscale medical imagery
(chest radiographs; brain scans), for which the domain-asymmetry indices EROS and PatchNN were
originally built and where these methods may fare better than they do here. The timings are single-machine CPU
medians at library-default thread settings, not a hardware-independent claim: on a GPU the
deep methods' constant shrinks substantially, whereas the classical methods need no
accelerator at all.

\section{Conclusions}\label{sec:conclusions}
Benchmarking thirteen symmetry scorers, from hand-crafted measures to frozen deep readouts,
under a chance-anchored, significance-tested protocol resolves mirror-symmetry discrimination
into a single property: it lives in mid-scale, unsigned, oriented features. Deep backbones peak
at a low or mid stage, a histogram-of-oriented-gradients descriptor peaks at a mid cell size, and
columnar Vision Transformers fall behind at coarse $16$-px tokenization but partly recover
with finer patches, placing the effect on
spatial structure rather than model capacity. The practical consequence is that a tuned classical
HOG comes within $0.03$ skill of the best frozen-network readout at ${\sim}300\times$ lower CPU
cost; among the methods tested, frozen deep features offer little over a tuned classical descriptor
for measuring symmetry. Whether scorers trained for the task can widen that margin is untested and
the natural next step. We release the scorers, the reflection-exact harness, and all analysis
scripts in \texttt{imgsym}, our open image-symmetry toolkit, for reuse and extension.

\vspace{6pt}

\funding{This research received no external funding.} %

\dataavailability{\texttt{imgsym} v0.1.0, an open library providing the thirteen scorers behind a
common interface, the reflection-exact evaluation harness, and the scripts to reproduce all
results, is available at \url{https://github.com/maxwoe/imgsym} (accessed on 1 August 2026). The
image datasets used are publicly available from their original sources (Section~\ref{sec:datasets}).}  %

\acknowledgments{During the preparation of this manuscript, the author used Claude Code (Anthropic, Claude Opus 4.x) %
 to assist with method implementation, experiment tooling, data analysis, and drafting and revising text passages. The author has reviewed and edited the output and takes full responsibility for the content of this publication. Open Access Funding by the University of Vienna.} %

\conflictsofinterest{The author is the owner of MAXI solutions e.U., a software company whose work includes computer-vision applications. The author declares no other conflicts of interest.} %

\newpage

\abbreviations{Abbreviations}{~%
The following abbreviations are used in this manuscript:\\

\noindent
\begin{tabular}{@{}ll}
AUC & Area under the ROC curve\\
CI & Confidence interval\\
CNN & Convolutional neural network\\
CPU & Central processing unit\\
DCT & Discrete cosine transform\\
GPU & Graphics processing unit\\
HOG & Histogram of oriented gradients\\
IQR & Interquartile range\\
JPEG & Joint Photographic Experts Group\\
PHOG & Pyramid histogram of oriented gradients\\
ROC & Receiver operating characteristic\\
SSIM & Structural similarity index measure\\
ViT & Vision Transformer\\
\end{tabular}
}

\appendixtitles{yes}
\appendixstart
\appendix
\section{The Backbone Pool of the Stage Ablation}\label{app:backbones}
Table~\ref{tab:backbones} lists the 17 pretrained \texttt{timm} backbones swept in the stage
ablation (Figure~\ref{fig:stage}), with each backbone's number of feature stages and its best
stage and skill on the easy (NYU + symComp17, 40 images each, $n=80$) and hard
(PIX2PER-art, $n=40$) subsets. The three columnar ViTs (ViT-Tiny/16, ViT-Base/16, and
ViT-Base/8; AugReg ImageNet-21k weights fine-tuned on ImageNet-1k) expose three
same-resolution feature stages; the patch-16 pair serves as the no-spatial-hierarchy control,
and ViT-Base/8 is the finer-tokenization control added to resolve the stride confound of
Section~\ref{sec:principle}.
The hard-subset point-estimate ordering should not be read as evidence for one hierarchical
backbone over another: at $n=40$ every interval half-width exceeds the largest
backbone-to-backbone gap, and the ordering is not preserved across subsets (MambaOut-Base,
seventh on the hard subset, co-leads the easy subset). The findings that are stable
across subsets are those of Section~\ref{sec:principle}: low/mid stages win, and the patch-16
columnar ViTs lose.

\begin{table}[H]
\caption{The 17-backbone pool of the stage ablation, sorted by hard-subset skill: number of
feature stages and the best stage with its skill and 95\% CI per subset.\label{tab:backbones}}
\begin{adjustwidth}{-\extralength}{0cm}
\begin{tabularx}{\fulllength}{Xccccc}
\toprule
 \multirow{2}{*}{\textbf{Backbone (\texttt{Timm} Identifier)}} &  \multirow{2}{*}{\textbf{Stages}} & \multicolumn{2}{c}{\textbf{Easy Subset}} & \multicolumn{2}{c}{\textbf{Hard Subset}}\\ %
 & & \textbf{Best} & \textbf{Skill [95\% CI]} & \textbf{Best} & \textbf{Skill [95\% CI]} \\
\midrule
\texttt{\footnotesize efficientnetv2\_rw\_t.ra2\_in1k} & 5 & 3 & +0.93 [+0.90, +0.96] & 3 & +0.73 [+0.62, +0.82]\\
\texttt{\footnotesize convnextv2\_atto.fcmae\_ft\_in1k} & 4 & 1 & +0.93 [+0.90, +0.96] & 1 & +0.72 [+0.62, +0.81]\\
\texttt{\footnotesize mambaout\_tiny.in1k} & 4 & 1 & +0.93 [+0.89, +0.97] & 0 & +0.72 [+0.60, +0.82]\\
\texttt{\footnotesize efficientvit\_b3.r224\_in1k} & 4 & 1 & +0.93 [+0.89, +0.96] & 3 & +0.72 [+0.63, +0.81]\\
\texttt{\footnotesize efficientvit\_b0.r224\_in1k} & 4 & 1 & +0.93 [+0.89, +0.96] & 3 & +0.71 [+0.60, +0.81]\\
\texttt{\footnotesize focalnet\_tiny\_lrf.ms\_in1k} & 4 & 1 & +0.93 [+0.88, +0.96] & 2 & +0.70 [+0.56, +0.82]\\
\texttt{\footnotesize mambaout\_base.in1k} & 4 & 1 & +0.94 [+0.90, +0.97] & 1 & +0.68 [+0.53, +0.80]\\
\texttt{\footnotesize efficientnetv2\_rw\_m.agc\_in1k} & 5 & 3 & +0.93 [+0.90, +0.96] & 3 & +0.68 [+0.55, +0.80]\\
\texttt{\footnotesize convnextv2\_base.fcmae\_ft\_in22k\_in1k} & 4 & 1 & +0.94 [+0.90, +0.97] & 1 & +0.67 [+0.51, +0.80]\\
\texttt{\footnotesize swin\_tiny\_patch4\_window7\_224} & 4 & 1 & +0.93 [+0.90, +0.96] & 1 & +0.67 [+0.54, +0.79]\\
\texttt{\footnotesize focalnet\_base\_lrf.ms\_in1k} & 4 & 1 & +0.93 [+0.89, +0.97] & 1 & +0.67 [+0.55, +0.78]\\
\texttt{\footnotesize swin\_base\_patch4\_window7\_224} & 4 & 1 & +0.93 [+0.90, +0.96] & 1 & +0.67 [+0.54, +0.79]\\
\texttt{\footnotesize edgenext\_xx\_small.in1k} & 4 & 1 & +0.92 [+0.88, +0.95] & 1 & +0.67 [+0.53, +0.78]\\
\texttt{\footnotesize vit\_base\_patch8\_224.augreg\_in21k\_ft\_in1k} & 3 & 0 & +0.92 [+0.89, +0.95] & 0 & +0.67 [+0.55, +0.79]\\
\texttt{\footnotesize edgenext\_base.in21k\_ft\_in1k} & 4 & 1 & +0.93 [+0.89, +0.96] & 3 & +0.64 [+0.51, +0.76]\\
\texttt{\footnotesize vit\_base\_patch16\_224.augreg\_in21k\_ft\_in1k} & 3 & 0 & +0.92 [+0.88, +0.95] & 0 & +0.61 [+0.46, +0.74]\\
\texttt{\footnotesize vit\_tiny\_patch16\_224.augreg\_in21k\_ft\_in1k} & 3 & 0 & +0.86 [+0.79, +0.93] & 0 & +0.39 [+0.20, +0.57]\\
\bottomrule
\end{tabularx}

\end{adjustwidth}
\end{table}

\section{Per-Dataset Results}\label{app:perdataset}
Table~\ref{tab:perdataset} reports every method's discrimination skill with its 95\%
bootstrap confidence interval on each of the nine datasets separately (the means of these
columns, with their own intervals, %
are reported in Table~\ref{tab:leaderboard}), so individual claims can
be checked and future methods compared against the benchmark per dataset.

\begin{table}[H]
\fontsize{8.7pt}{8.7pt}\selectfont
\caption{Per-dataset discrimination skill with 95\% bootstrap confidence intervals, under the
single-axis (top) and multi-axis (bottom) protocols; rows sorted by mean single-axis skill;
bold as in Table~\ref{tab:leaderboard}.\label{tab:perdataset}}
\setlength{\cellWidtha}{\fulllength/6-2\tabcolsep+0.05in}
\setlength{\cellWidthb}{\fulllength/6-2\tabcolsep-0.01in}
\setlength{\cellWidthc}{\fulllength/6-2\tabcolsep-0.01in}
\setlength{\cellWidthd}{\fulllength/6-2\tabcolsep-0.01in}
\setlength{\cellWidthe}{\fulllength/6-2\tabcolsep-0.01in}
\setlength{\cellWidthf}{\fulllength/6-2\tabcolsep-0.01in}
 \begin{adjustwidth}{-\extralength}{0cm}
  \begin{tabularx}{\fulllength}{>{\raggedright\arraybackslash}m{\cellWidtha}>{\centering\arraybackslash}m{\cellWidthb}>{\centering\arraybackslash}m{\cellWidthc}>{\centering\arraybackslash}m{\cellWidthd}>{\centering\arraybackslash}m{\cellWidthe}>{\centering\arraybackslash}m{\cellWidthf}}
   \toprule
\textbf{Single-Axis Protocol} & \textbf{symComp17} & \multicolumn{2}{c}{\hspace{-40pt}\textbf{NYU}} & \textbf{\hspace{-60pt}PIX2PER-nat} & \textbf{PIX2PER-art}\\\midrule

\textbf{DeepFeat} & +0.87 [+0.82, +0.91] & \multicolumn{2}{c}{\hspace{-40pt}+0.94 [+0.92, +0.95]} & \hspace{-60pt}+0.77 [+0.73, +0.82] & +0.74 [+0.69, +0.78]\\
AlexNet-C2 & +0.86 [+0.82, +0.90] & \multicolumn{2}{c}{\hspace{-40pt}+0.91 [+0.89, +0.94]} & \hspace{-60pt}+0.77 [+0.73, +0.81] & +0.71 [+0.66, +0.76]\\
\textbf{HOG} & +0.86 [+0.82, +0.90] & \multicolumn{2}{c}{\hspace{-40pt}+0.92 [+0.90, +0.95]} & \hspace{-60pt}+0.69 [+0.64, +0.75] & +0.73 [+0.67, +0.77]\\
Gabor & +0.76 [+0.69, +0.81] & \multicolumn{2}{c}{\hspace{-40pt}+0.83 [+0.78, +0.87]} & \hspace{-60pt}+0.64 [+0.58, +0.69] & +0.61 [+0.56, +0.67]\\
Grad & +0.74 [+0.66, +0.80] & \multicolumn{2}{c}{\hspace{-40pt}+0.83 [+0.77, +0.87]} & \hspace{-60pt}+0.61 [+0.55, +0.68] & +0.51 [+0.43, +0.57]\\
MS-Grad & +0.72 [+0.64, +0.79] & \multicolumn{2}{c}{\hspace{-40pt}+0.81 [+0.76, +0.85]} & \hspace{-60pt}+0.55 [+0.49, +0.62] & +0.51 [+0.44, +0.59]\\
PatchNN & +0.70 [+0.63, +0.78] & \multicolumn{2}{c}{\hspace{-40pt}+0.78 [+0.74, +0.82]} & \hspace{-60pt}+0.59 [+0.52, +0.65] & +0.51 [+0.45, +0.57]\\
SlideWin & +0.71 [+0.63, +0.79] & \multicolumn{2}{c}{\hspace{-40pt}+0.77 [+0.72, +0.82]} & \hspace{-60pt}+0.54 [+0.48, +0.61] & +0.55 [+0.49, +0.61]\\
PixCorr & +0.69 [+0.61, +0.77] & \multicolumn{2}{c}{\hspace{-40pt}+0.77 [+0.72, +0.81]} & \hspace{-60pt}+0.54 [+0.48, +0.61] & +0.55 [+0.49, +0.61]\\
PHOG & +0.63 [+0.55, +0.71] & \multicolumn{2}{c}{\hspace{-40pt}+0.75 [+0.70, +0.80]} & \hspace{-60pt}+0.58 [+0.52, +0.64] & +0.54 [+0.48, +0.61]\\
DCT & +0.66 [+0.57, +0.74] & \multicolumn{2}{c}{\hspace{-40pt}+0.73 [+0.67, +0.78]} & \hspace{-60pt}+0.51 [+0.44, +0.58] & +0.53 [+0.46, +0.59]\\
WBS & +0.67 [+0.59, +0.75] & \multicolumn{2}{c}{\hspace{-40pt}+0.72 [+0.67, +0.78]} & \hspace{-60pt}+0.46 [+0.40, +0.53] & +0.45 [+0.39, +0.52]\\
EROS & +0.43 [+0.34, +0.52] & \multicolumn{2}{c}{\hspace{-40pt}+0.46 [+0.39, +0.53]} & \hspace{-60pt}+0.17 [+0.09, +0.25] & +0.04 [$-$0.06, +0.12]\\\midrule

\textbf{Multi-axis protocol} & \textbf{symComp13} & \textbf{NYU} & \textbf{DENDI} & \textbf{PIX2PER-nat} & \textbf{PIX2PER-art}\\
\midrule  %

\textbf{DeepFeat} & +0.82 [+0.75, +0.88] & +0.87 [+0.83, +0.90] & +0.52 [+0.50, +0.54] & +0.63 [+0.59, +0.66] & +0.65 [+0.62, +0.68]\\
AlexNet-C2 & +0.74 [+0.65, +0.84] & +0.78 [+0.73, +0.83] & +0.41 [+0.39, +0.43] & +0.61 [+0.57, +0.64] & +0.63 [+0.60, +0.66]\\
\textbf{HOG} & +0.79 [+0.72, +0.87] & +0.80 [+0.75, +0.85] & +0.49 [+0.47, +0.52] & +0.57 [+0.54, +0.60] & +0.60 [+0.56, +0.63]\\
Gabor & +0.63 [+0.48, +0.77] & +0.72 [+0.67, +0.77] & +0.27 [+0.25, +0.29] & +0.51 [+0.48, +0.55] & +0.53 [+0.50, +0.56]\\
Grad & +0.50 [+0.35, +0.66] & +0.72 [+0.66, +0.78] & +0.39 [+0.36, +0.41] & +0.42 [+0.38, +0.46] & +0.45 [+0.42, +0.49]\\
MS-Grad & +0.52 [+0.37, +0.67] & +0.62 [+0.56, +0.69] & +0.31 [+0.28, +0.33] & +0.40 [+0.35, +0.44] & +0.39 [+0.35, +0.43]\\
PatchNN & +0.59 [+0.46, +0.71] & +0.59 [+0.52, +0.65] & +0.24 [+0.22, +0.26] & +0.38 [+0.35, +0.42] & +0.44 [+0.40, +0.47]\\
SlideWin & +0.53 [+0.36, +0.69] & +0.64 [+0.57, +0.70] & +0.29 [+0.27, +0.32] & +0.41 [+0.38, +0.45] & +0.49 [+0.46, +0.52]\\
PixCorr & +0.57 [+0.43, +0.70] & +0.63 [+0.57, +0.69] & +0.28 [+0.26, +0.31] & +0.43 [+0.39, +0.46] & +0.51 [+0.48, +0.55]\\
PHOG & +0.55 [+0.43, +0.66] & +0.62 [+0.56, +0.68] & +0.33 [+0.30, +0.35] & +0.48 [+0.45, +0.51] & +0.47 [+0.44, +0.51]\\
DCT & +0.45 [+0.29, +0.60] & +0.56 [+0.50, +0.63] & +0.24 [+0.22, +0.27] & +0.39 [+0.35, +0.43] & +0.47 [+0.44, +0.51]\\
WBS & +0.46 [+0.31, +0.61] & +0.61 [+0.55, +0.66] & +0.24 [+0.22, +0.27] & +0.33 [+0.29, +0.36] & +0.38 [+0.35, +0.42]\\
EROS & +0.27 [+0.13, +0.40] & +0.28 [+0.20, +0.36] & +0.16 [+0.14, +0.18] & +0.00 [$-$0.04, +0.05] & +0.07 [+0.03, +0.12]\\
   \bottomrule
  \end{tabularx}
 \end{adjustwidth}

\end{table}

\begin{adjustwidth}{-\extralength}{0cm}
\reftitle{References} %

\PublishersNote{}
\end{adjustwidth}


\begin{thebibliography}{999}

\bibitem[Reisfeld et~al.(1990)Reisfeld, Wolfson, and
  Yeshurun]{reisfeldDetectionInterestPoints1990}
Reisfeld, D.; Wolfson, H.; Yeshurun, Y.
\newblock Detection of Interest Points Using Symmetry.
\newblock In Proceedings of the Third {{International Conference}} on
  {{Computer Vision}}, Osaka, Japan,  4--7 December {1990}; pp. 62--65.
\newblock {\url{https://doi.org/10.1109/ICCV.1990.139494}}. %


\bibitem[Bonneh et~al.(2003)Bonneh, Reisfeld, and
  Yeshurun]{bonneh2003quantification}
Bonneh, Y.; Reisfeld, D.; Yeshurun, Y.
\newblock Quantification of Local Symmetry: Application to Texture
  Discrimination. In {\em Human Symmetry Perception and Its Computational
  Analysis}; Psychology Press: Mahwah, NJ, USA, {2003}; pp. 310--325.  %



\bibitem[Funk and Liu(2016)]{funkSymmetryReCAPTCHA2016}
Funk, C.; Liu, Y.
\newblock Symmetry {{reCAPTCHA}}.
\newblock In Proceedings of the 2016 {{IEEE Conference}} on {{Computer Vision}}
  and {{Pattern Recognition}} ({{CVPR}}), Las Vegas, NV, USA,  27--30 June {2016}; pp. 5165--5174.
\newblock {\url{https://doi.org/10.1109/CVPR.2016.558}}. %

\bibitem[Smith and Jenkinson(1999)]{smithAccurateRobustSymmetry1999}
Smith, S.; Jenkinson, M.
\newblock Accurate {{Robust Symmetry Estimation}}. In {\em Medical {{Image
  Computing}} and {{Computer-Assisted Intervention}}---{{MICCAI}}'99}; Goos,
  G., Hartmanis, J., Van~Leeuwen, J., Taylor, C., Colchester, A., Eds.;
  Springer: Berlin/Heidelberg, Germany, 1999; Volume 1679, pp.
  308--317.
\newblock {\url{https://doi.org/10.1007/10704282_34}}.

\bibitem[Hogeweg et~al.(2017)Hogeweg, S{\'a}nchez, Maduskar, Philipsen, and
  Van~Ginneken]{hogewegFastEffectiveQuantification2017}
Hogeweg, L.; S{\'a}nchez, C.I.; Maduskar, P.; Philipsen, R.H.; Van~Ginneken, B.
\newblock Fast and Effective Quantification of Symmetry in Medical Images for
  Pathology Detection: {{Application}} to Chest Radiography.
\newblock {\em Med. Phys.} {\bf 2017}, {\em 44},~2242--2256.
\newblock {\url{https://doi.org/10.1002/mp.12127}}.

\bibitem[Frey et~al.(2007)Frey, Robertson, and
  Bukoski]{freyMethodQuantifyingRotational2007}
Frey, F.M.; Robertson, A.; Bukoski, M.
\newblock A Method for Quantifying Rotational Symmetry.
\newblock {\em New Phytol.} {\bf 2007}, {\em 175},~785--791.
\newblock {\url{https://doi.org/10.1111/j.1469-8137.2007.02146.x}}.

\bibitem[Mayer and Landwehr(2018)]{mayerQuantifyingVisualAesthetics2018}
Mayer, S.; Landwehr, J.R.
\newblock Quantifying Visual Aesthetics Based on Processing Fluency Theory:
  {{Four}} Algorithmic Measures for Antecedents of Aesthetic Preferences.
\newblock {\em Psychol. Aesthet. Creat. Arts} {\bf 2018},
  {\em 12},~399--431.
\newblock {\url{https://doi.org/10.1037/aca0000187}}.

\bibitem[Gartus and Leder(2017)]{gartusPredictingPerceivedVisual2017}
Gartus, A.; Leder, H.
\newblock Predicting Perceived Visual Complexity of Abstract Patterns Using
  Computational Measures: {{The}} Influence of Mirror Symmetry on Complexity
  Perception.
\newblock {\em PLoS ONE} {\bf 2017}, {\em 12},~e0185276.
\newblock {\url{https://doi.org/10.1371/journal.pone.0185276}}.
\newpage
\bibitem[Gnutti et~al.(2021)Gnutti, Guerrini, and
  Leonardi]{gnuttiCombiningAppearanceGradient2021}
Gnutti, A.; Guerrini, F.; Leonardi, R.
\newblock Combining {{Appearance}} and {{Gradient Information}} for {{Image
  Symmetry Detection}}.
\newblock {\em IEEE Trans. Image Process.} {\bf 2021}, {\em 30},~5708--5723.
\newblock {\url{https://doi.org/10.1109/TIP.2021.3085202}}.

\bibitem[Shaker and Monadjemi(2015)]{shakerNewSymmetryMeasure2015}
Shaker, F.; Monadjemi, A.
\newblock A New Symmetry Measure Based on {{Gabor}} Filters.
\newblock In Proceedings of the 2015 23rd {{Iranian Conference}} on
  {{Electrical Engineering}}, Tehran, Iran,  10--14 May {2015}; pp. 705--710.
\newblock {\url{https://doi.org/10.1109/IranianCEE.2015.7146305}}. %

\bibitem[Gunlu and Bilge(2009)]{gunluSymmetryAnalysis2D2009}
Gunlu, G.; Bilge, H.S.
\newblock Symmetry Analysis for {{2D}} Images by Using {{DCT}} Coefficients.
\newblock In Proceedings of the 2009 {{Fifth International Conference}} on
  {{Soft Computing}}, {{Computing}} with {{Words}} and {{Perceptions}} in
  {{System Analysis}}, {{Decision}} and {{Control}}, Famagusta, Cyprus,  2--4 September {2009}; pp. 1--4.
\newblock {\url{https://doi.org/10.1109/ICSCCW.2009.5379480}}. %

\bibitem[Brachmann and Redies(2016)]{brachmannUsingConvolutionalNeural2016}
Brachmann, A.; Redies, C.
\newblock Using {{Convolutional Neural Network Filters}} to {{Measure
  Left-Right Mirror Symmetry}} in {{Images}}.
\newblock {\em Symmetry} {\bf 2016}, {\em 8},~144.
\newblock {\url{https://doi.org/10.3390/sym8120144}}.

\bibitem[Liu et~al.(2013)Liu, Slota, Zheng, Wu, Park, Lee, Rauschert, and
  Liu]{liuSymmetryDetectionRealWorld2013}
Liu, J.; Slota, G.; Zheng, G.; Wu, Z.; Park, M.; Lee, S.; Rauschert, I.; Liu,
  Y.
\newblock Symmetry {{Detection}} from {{RealWorld Images Competition}} 2013:
  {{Summary}} and {{Results}}.
\newblock In Proceedings of the 2013 {{IEEE Conference}} on {{Computer Vision}}
  and {{Pattern Recognition Workshops}}, Portland, OR, USA,  23--28 June {2013}; pp. 200--205. %
\newblock {\url{https://doi.org/10.1109/CVPRW.2013.155}}.

\bibitem[Funk et~al.(2017)Funk, Lee, Oswald, Tsogkas, Shen, Cohen, Dickinson,
  and Liu]{funk2017ICCVChallenge2017a}
Funk, C.; Lee, S.; Oswald, M.R.; Tsogkas, S.; Shen, W.; Cohen, A.; Dickinson,
  S.; Liu, Y.
\newblock 2017 {{ICCV Challenge}}: {{Detecting Symmetry}} in the {{Wild}}.
\newblock In Proceedings of the 2017 {{IEEE International Conference}} on
  {{Computer Vision Workshops}} ({{ICCVW}}), Venice,  Italy, 22--29 October {2017}; pp. 1692--1701. %
\newblock {\url{https://doi.org/10.1109/ICCVW.2017.198}}.

\bibitem[Loy and Eklundh(2006)]{loyDetectingSymmetrySymmetric2006}
Loy, G.; Eklundh, J.O.
\newblock Detecting {{Symmetry}} and {{Symmetric Constellations}} of
  {{Features}}. In {\em Computer {{Vision}}---{{ECCV}} 2006}; Leonardis, A.,
  Bischof, H., Pinz, A., Eds.; Springer: Berlin/Heidelberg, Germany, 
  2006; Volume 3952, pp. 508--521.
\newblock {\url{https://doi.org/10.1007/11744047_39}}.

\bibitem[Cho and Lee(2009)]{choBilateralSymmetryDetection}
Cho, M.; Lee, K.M.
\newblock Bilateral {{Symmetry Detection}} via {{Symmetry-Growing}}.
\newblock In Proceedings of the British {{Machine Vision Conference}} 2009,
  London, UK,  7--10 September {2009}; pp. 1--11.
\newblock {\url{https://doi.org/10.5244/C.23.4}}. %

\bibitem[Cai et~al.(2014)Cai, Li, Su, and
  Zhao]{caiAdaptiveSymmetryDetection2014}
Cai, D.; Li, P.; Su, F.; Zhao, Z.
\newblock An Adaptive Symmetry Detection Algorithm Based on Local Features.
\newblock In Proceedings of the 2014 {{IEEE Visual Communications}} and {{Image
  Processing Conference}}, Valletta, Malta,  7--10 December {2014}; pp. 478--481.
\newblock {\url{https://doi.org/10.1109/VCIP.2014.7051610}}. %

\bibitem[Wang et~al.(2015)Wang, Tang, and
  Zhang]{wangReflectionSymmetryDetection2015}
Wang, Z.; Tang, Z.; Zhang, X.
\newblock Reflection {{Symmetry Detection Using Locally Affine Invariant Edge
  Correspondence}}.
\newblock {\em IEEE Trans. Image Process.} {\bf 2015}, {\em 24},~1297--1301.
\newblock {\url{https://doi.org/10.1109/TIP.2015.2393060}}.

\bibitem[Sun and Si(1999)]{sunFastReflectionalSymmetry1999}
Sun, C.; Si, D.
\newblock Fast {{Reflectional Symmetry Detection Using Orientation
  Histograms}}.
\newblock {\em Real-Time Imaging} {\bf 1999}, {\em 5},~63--74.
\newblock {\url{https://doi.org/10.1006/rtim.1998.0135}}.

\bibitem[Cicconet et~al.(2016)Cicconet, Birodkar, Lund, Werman, and
  Geiger]{cicconetConvolutionalApproachReflection2016}
Cicconet, M.; Birodkar, V.; Lund, M.; Werman, M.; Geiger, D.
\newblock A Convolutional Approach to Reflection Symmetry.
\textit{arXiv} \textbf{2016}, arXiv:1609.05257.
\newblock {\url{https://doi.org/10.48550/arXiv.1609.05257}}. %

\bibitem[Elawady et~al.(2017)Elawady, Ducottet, Alata, Barat, and
  Colantoni]{elawadyWaveletBasedReflectionSymmetry2017}
Elawady, M.; Ducottet, C.; Alata, O.; Barat, C.; Colantoni, P.
\newblock Wavelet-{{Based Reflection Symmetry Detection}} via {{Textural}} and
  {{Color Histograms}}.
\newblock In Proceedings of the 2017 {{IEEE International Conference}} on
  {{Computer Vision Workshops}} ({{ICCVW}}), Venice, Italy,  22--29 October {2017}; pp. 1725--1733. %
\newblock {\url{https://doi.org/10.1109/ICCVW.2017.202}}.

\bibitem[Guerrini et~al.(2017)Guerrini, Gnutti, and
  Leonardi]{guerriniImageSymmetriesRight2017}
Guerrini, F.; Gnutti, A.; Leonardi, R.
\newblock Image Symmetries: {{The}} Right Balance between Evenness and
  Perception.
\newblock In Proceedings of the 2017 {{International Conference}} on
  {{Systems}}, {{Signals}} and {{Image Processing}} ({{IWSSIP}}), Pozna\'n,
  Poland,  22--24 May {2017}; pp. 1--5. %
\newblock {\url{https://doi.org/10.1109/IWSSIP.2017.7965605}}.

\bibitem[Xiao et~al.(2005)Xiao, Hou, Miao, and
  Wang]{xiaoUsingPhaseInformation2005}
Xiao, Z.; Hou, Z.; Miao, C.; Wang, J.
\newblock Using Phase Information for Symmetry Detection.
\newblock {\em Pattern Recognit. Lett.} {\bf 2005}, {\em 26},~1985--1994.
\newblock {\url{https://doi.org/10.1016/j.patrec.2005.02.003}}.

\bibitem[Cicconet et~al.(2017)Cicconet, Hildebrand, and
  Elliott]{cicconetFindingMirrorSymmetry2017}
Cicconet, M.; Hildebrand, D.G.C.; Elliott, H.
\newblock Finding {{Mirror Symmetry}} via {{Registration}} and {{Optimal
  Symmetric Pairwise Assignment}} of {{Curves}}.
\newblock In Proceedings of the 2017 {{IEEE International Conference}} on
  {{Computer Vision Workshops}} ({{ICCVW}}), Venice,  Italy, 22--29 October {2017}; pp. 1749--1758.  %
\newblock {\url{https://doi.org/10.1109/ICCVW.2017.206}}.

\bibitem[Tsogkas and Kokkinos(2012)]{tsogkasLearningBasedSymmetryDetection2012}
Tsogkas, S.; Kokkinos, I.
\newblock Learning-{{Based Symmetry Detection}} in {{Natural Images}}. In {\em
  Computer {{Vision}}---{{ECCV}} 2012}; Hutchison, D., Kanade, T., Kittler,
  J., Kleinberg, J.M., Mattern, F., Mitchell, J.C., Naor, M., Nierstrasz, O.,
  Pandu~Rangan, C., Steffen, B.,  et~al., Eds.; Springer:
  Berlin/Heidelberg, Germany, 2012; Volume 7578, pp. 41--54.
\newblock {\url{https://doi.org/10.1007/978-3-642-33786-4_4}}.

\bibitem[Ke et~al.(2017)Ke, Chen, Jiao, Zhao, and
  Ye]{keSRNSideOutputResidual2017}
Ke, W.; Chen, J.; Jiao, J.; Zhao, G.; Ye, Q.
\newblock {{SRN}}: {{Side-Output Residual Network}} for {{Object Symmetry
  Detection}} in the {{Wild}}.
\newblock In Proceedings of the 2017 {{IEEE Conference}} on {{Computer Vision}}
  and {{Pattern Recognition}} ({{CVPR}}), Honolulu, HI, USA, 21--26 July {2017}; pp. 302--310. %
\newblock {\url{https://doi.org/10.1109/CVPR.2017.40}}.

\bibitem[Seo et~al.(2022)Seo, Kim, Kwak, and
  Cho]{seoReflectionRotationSymmetry2022}
Seo, A.; Kim, B.; Kwak, S.; Cho, M.
\newblock Reflection and {{Rotation Symmetry Detection}} via {{Equivariant
  Learning}}.
\newblock In Proceedings of the 2022 {{IEEE}}/{{CVF Conference}} on {{Computer
  Vision}} and {{Pattern Recognition}} ({{CVPR}}), New Orleans, LA, USA,  19--24 June {2022}; \mbox{pp. 9529--9538.} %
\newblock {\url{https://doi.org/10.1109/CVPR52688.2022.00932}}.

\bibitem[Yu et~al.(2025)Yu, Seo, and Cho]{yuAxislevelSymmetryDetection2025}
Yu, W.; Seo, A.; Cho, M.
\newblock Axis-Level {{Symmetry Detection}} with {{Group-Equivariant
  Representation}}. arXiv \textbf{2025}, arXiv:2508.10740.
\newblock {\url{https://doi.org/10.48550/arXiv.2508.10740}}. %

\bibitem[Yang et~al.(2025)Yang, Rahman, and
  Yeh]{yangCLIPSymDelvingSymmetry2025}
Yang, T.; Rahman, M.A.; Yeh, R.A.
\newblock {{CLIPSym}}: {{Delving}} into {{Symmetry Detection}} with {{CLIP}}.
arXiv \textbf{2025}, arXiv:2508.14197.
\newblock {\url{https://doi.org/10.48550/arXiv.2508.14197}}. %

\bibitem[Nguyen(2019)]{nguyenProjectionBasedApproach2019}
Nguyen, T.P.
\newblock Projection {{Based Approach}} for {{Reflection Symmetry Detection}}.
\newblock In Proceedings of the 2019 {{IEEE International Conference}} on
  {{Image Processing}} ({{ICIP}}), Taipei, Taiwan,  22--25 September {2019}; pp. 4235--4239.  %
\newblock {\url{https://doi.org/10.1109/ICIP.2019.8803575}}.

\bibitem[Nguyen et~al.(2022)Nguyen, Truong, Nguyen, and
  Kim]{nguyenReflectionSymmetryDetection2022}
Nguyen, T.P.; Truong, H.P.; Nguyen, T.T.; Kim, Y.G.
\newblock Reflection Symmetry Detection of Shapes Based on Shape Signatures.
\newblock {\em Pattern Recognit.} {\bf 2022}, {\em 128},~108667.
\newblock {\url{https://doi.org/10.1016/j.patcog.2022.108667}}.

\bibitem[Lomov et~al.(2022)Lomov, Seredin, and
  Kushnir]{lomovDetectionOptimalReflection2022}
Lomov, N.; Seredin, O.; Kushnir, O.
\newblock Detection of the {{Optimal Reflection Symmetry Axis}} with the
  {{Jaccard Index}} and the {{Radon Transform}}.
\newblock In Proceedings of the 2022 {{International Russian Automation
  Conference}} ({{RusAutoCon}}), Sochi, Russia,  4--10 September {2022}; pp.
  489--498. %
\newblock {\url{https://doi.org/10.1109/RusAutoCon54946.2022.9896373}}.

\bibitem[Mestetskiy and
  Zhuravskaya(2020)]{mestetskiyMirrorSymmetryDetection2020}
Mestetskiy, L.; Zhuravskaya, A.
\newblock Mirror {{Symmetry Detection}} in {{Digital Images}}.
\newblock In Proceedings of the 15th {{International Joint Conference}} on
  {{Computer Vision}}, {{Imaging}} and {{Computer Graphics Theory}} and
  {{Applications}}, Valletta, Malta,  27--29 February {2020}; pp. 331--337.
\newblock {\url{https://doi.org/10.5220/0008976003310337}}.  %

\bibitem[Kushnir et~al.(2019)Kushnir, Seredin, and
  Fedotova]{kushnirALGORITHMSADJUSTMENTSYMMETRY2019}
Kushnir, O.A.; Seredin, O.S.; Fedotova, S.A.
\newblock Algorithms for Adjustment of Symmetry Axis Found for {{2D}} Shapes by
  the Skeleton Comparison {Method}.
\newblock {\em Int. Arch. Photogramm. Remote Sens. Spat. Inf. Sci.} {\bf
  2019}, {\em XLII-2/W12},~129--136.
\newblock {\url{https://doi.org/10.5194/isprs-archives-XLII-2-W12-129-2019}}.

\bibitem[Patraucean et~al.(2013)Patraucean, Von~Gioi, and
  Ovsjanikov]{patrauceanDetectionMirrorSymmetricImage2013}
Patraucean, V.; Von~Gioi, R.G.; Ovsjanikov, M.
\newblock Detection of {{Mirror-Symmetric Image Patches}}.
\newblock In Proceedings of the 2013 {{IEEE Conference}} on {{Computer Vision}}
  and {{Pattern Recognition Workshops}}, Portland, OR, USA,  23--28 June {2013}; pp. 211--216. %
\newblock {\url{https://doi.org/10.1109/CVPRW.2013.38}}.

\bibitem[Von~Gioi and Patraucean(2015)]{vongioiContrarioPatchMatching2015}
Von~Gioi, R.G.; Patraucean, V.
\newblock A Contrario Patch Matching, with an Application to Keypoint Matches
  Validation.
\newblock In Proceedings of the 2015 {{IEEE International Conference}} on
  {{Image Processing}} ({{ICIP}}), Quebec City, QC, Canada,  27--30 September  {2015}; \mbox{pp. 946--950}. %
\newblock {\url{https://doi.org/10.1109/ICIP.2015.7350939}}.

\bibitem[Dalitz et~al.(2019)Dalitz, Wilberg, and
  Jeltsch]{dalitzGradientProductTransform2019}
Dalitz, C.; Wilberg, J.; Jeltsch, M.
\newblock The {{Gradient Product Transform}}: {{An Image Filter}} for
  {{Symmetry Detection}}.
\newblock {\em Image Process. Line} {\bf 2019}, {\em 9},~413--431.
\newblock {\url{https://doi.org/10.5201/ipol.2019.270}}.

\bibitem[Bauerly and Liu(2006)]{bauerlyComputationalModelingExperimental2006}
Bauerly, M.; Liu, Y.
\newblock Computational Modeling and Experimental Investigation of Effects of
  Compositional Elements on Interface and Design Aesthetics.
\newblock {\em Int. J. Hum.-Comput. Stud.} {\bf 2006},
  {\em 64},~670--682.
\newblock {\url{https://doi.org/10.1016/j.ijhcs.2006.01.002}}.

\bibitem[H{\"u}bner and Fillinger(2016)]{hubnerComparisonObjectiveMeasures2016}
H{\"u}bner, R.; Fillinger, M.G.
\newblock Comparison of {{Objective Measures}} for {{Predicting Perceptual
  Balance}} and {{Visual Aesthetic Preference}}.
\newblock {\em Front. Psychol.} {\bf 2016}, {\em 7},~335.
\newblock {\url{https://doi.org/10.3389/fpsyg.2016.00335}}.

\bibitem[Liu et~al.(2000)Liu, Collins, and
  Rothfus]{liuRobustMidsagittalPlane2000}
Liu, Y.; Collins, R.T.; Rothfus, W.E.
\newblock Robust {{Midsagittal Plane Extraction}} from {{Coarse}},
  {{Pathological 3D Images}}. In {\em Medical {{Image Computing}} and
  {{Computer-Assisted Intervention}}---{{MICCAI}} 2000}; Goos, G., Hartmanis,
  J., Van~Leeuwen, J., Delp, S.L., DiGoia, A.M., Jaramaz, B., Eds.; Springer: Berlin/Heidelberg,  Germany, 2000; Volume 1935, pp. 83--94.
\newblock {\url{https://doi.org/10.1007/978-3-540-40899-4_9}}.

\bibitem[{Renero-C} et~al.(2017){Renero-C}, {Romero-H}, and
  {Peregrina-B}]{renero-cExtractingSymmetryHuman2017}
{Renero-C}, F.J.; {Romero-H}, R.A.; {Peregrina-B}, H.
\newblock Extracting the Symmetry of the Human Face from Digital Photographs.
\newblock {\em Bio-Algorithms Med-Syst.} {\bf 2017}, {\em 13},~103--109.
\newblock {\url{https://doi.org/10.1515/bams-2017-0002}}.

\bibitem[Redies et~al.(2025)Redies, Bartho, Ko{\ss}mann, Spehar, H{\"u}bner,
  Wagemans, and {Hayn-Leichsenring}]{rediesToolboxCalculatingQuantitative2025}
Redies, C.; Bartho, R.; Ko{\ss}mann, L.; Spehar, B.; H{\"u}bner, R.; Wagemans,
  J.; {Hayn-Leichsenring}, G.U.
\newblock A Toolbox for Calculating Quantitative Image Properties in Aesthetics
  Research.
\newblock {\em Behav. Res. Methods} {\bf 2025}, {\em 57},~117.
\newblock {\url{https://doi.org/10.3758/s13428-025-02632-3}}.

\bibitem[Cui and Allen(2008)]{cuiImageQualityMetric2008}
Cui, L.; Allen, A.
\newblock An {{Image Quality Metric}} Based on {{Corner}}, {{Edge}} and
  {{Symmetry Maps}}.
\newblock In Proceedings of the British {{Machine Vision Conference}} 2008,
  Leeds,  UK, 1--4 September {2008}; pp. 38.1--38.10.
\newblock {\url{https://doi.org/10.5244/C.22.38}}. %

\bibitem[Nagar and Raman(2017)]{nagarSymmMapEstimation2D2017}
Nagar, R.; Raman, S.
\newblock {{SymmMap}}: {{Estimation}} of the 2-{{D Reflection Symmetry Map}}
  and {{Its Applications}}.
\newblock In Proceedings of the 2017 {{IEEE International Conference}} on
  {{Computer Vision Workshops}} ({{ICCVW}}), Venice, Italy,  22--29 October  {2017}; pp. 1715--1724. %
\newblock {\url{https://doi.org/10.1109/ICCVW.2017.201}}.

\bibitem[Murad{\'a}s~Odriozola et~al.(2026)Murad{\'a}s~Odriozola, Ko{\ss}mann,
  Tuytelaars, and Wagemans]{muradasodriozolaPixelsPerceptionBenchmark2026}
Murad{\'a}s~Odriozola, G.; Ko{\ss}mann, L.; Tuytelaars, T.; Wagemans, J.
\newblock From Pixels to Perception: {{A}} Benchmark for Human-like Symmetry
  Detection.
\newblock {\em Vis. Res.} {\bf 2026}, {\em 245},~108825.
\newblock {\url{https://doi.org/10.1016/j.visres.2026.108825}}.

\bibitem[Kiryati and Gofman(1998)]{kiryati1998detecting}
Kiryati, N.; Gofman, Y.
\newblock Detecting Symmetry in Grey Level Images: {{The}} Global Optimization
  Approach.
\newblock {\em Int. J. Comput. Vis.} {\bf 1998}, {\em
  29},~29--45.
\newblock {\url{https://doi.org/10.1023/A:1008034529558}}.

\bibitem[Dalal and Triggs(2005)]{dalalHistogramsOrientedGradients2005}
Dalal, N.; Triggs, B.
\newblock Histograms of {{Oriented Gradients}} for {{Human Detection}}.
\newblock In Proceedings of the 2005 {{IEEE Computer Society Conference}} on
  {{Computer Vision}} and {{Pattern Recognition}} ({{CVPR}}'05), San Diego, CA,
  USA,  20--26 June {2005}; Volume~1, \mbox{pp. 886--893}. 
\newblock {\url{https://doi.org/10.1109/CVPR.2005.177}}. %

\bibitem[Bosch et~al.(2007)Bosch, Zisserman, and
  Munoz]{boschRepresentingShapeSpatial2007}
Bosch, A.; Zisserman, A.; Munoz, X.
\newblock Representing Shape with a Spatial Pyramid Kernel.
\newblock In Proceedings of the 6th {{ACM}} International Conference on
  {{Image}} and Video Retrieval, Amsterdam, The Netherlands,  9--11 July  {2007}; pp.
  401--408. %
\newblock {\url{https://doi.org/10.1145/1282280.1282340}}.


\bibitem[Yu and Wang(2025)]{yuMambaOutWeReally}
Yu, W.; Wang, X.
\newblock {{MambaOut}}: {{Do We Really Need Mamba}} for {{Vision}}?
\newblock In Proceedings of the {{IEEE}}/{{CVF Conference}} on {{Computer
  Vision}} and {{Pattern Recognition}} ({{CVPR}}), Nashville, TN, USA,  11--15 June {2025}; pp. 4484--4496. 
\newblock \url{https://doi.org/10.1109/CVPR52734.2025.00423}. %

\bibitem[Wightman(2019)]{rw2019timm}
Wightman, R.
\newblock {{PyTorch}} Image Models. GitHub, {2019}. Available online: \url{https://github.com/huggingface/pytorch-image-models} (accessed on 1 August 2026). \url{https://doi.org/10.5281/zenodo.4414861}.




\bibitem[Bonneh and Tyler(2025)]{bonnehCanonicalDeepNeural2025}
Bonneh, Y.S.; Tyler, C.W.
\newblock The Canonical Deep Neural Network as a Model for Human Symmetry
  Processing.
\newblock {\em iScience} {\bf 2025}, {\em 28},~111540.
\newblock {\url{https://doi.org/10.1016/j.isci.2024.111540}}.

\bibitem[Kanezaki et~al.(2014)Kanezaki, Mukuta, and
  Harada]{kanezakiMirrorReflectionInvariant2014}
Kanezaki, A.; Mukuta, Y.; Harada, T.
\newblock Mirror Reflection Invariant {{HOG}} Descriptors for Object Detection.
\newblock In Proceedings of the 2014 {{IEEE International Conference}} on
  {{Image Processing}} ({{ICIP}}), Paris, France,  27--30 October {2014}; pp. 1594--1598. %
\newblock {\url{https://doi.org/10.1109/ICIP.2014.7025319}}.

\end{thebibliography}
\end{document}